\documentclass[journal]{IEEEtran}
\pdfoutput=1
\usepackage{lineno,latexsym,amsthm,amssymb,caption,color}
\usepackage[pdftex]{graphicx}
\usepackage[caption=false,font=footnotesize]{subfig}
\usepackage[fleqn]{amsmath}
\usepackage{algpseudocode}
\usepackage{algorithmicx,algorithm}
\usepackage{cite}
\usepackage{url}
\usepackage{hyperref}
\usepackage{array,booktabs}
\usepackage{multirow}
\usepackage{threeparttable}
\UseRawInputEncoding
\hypersetup{pdfauthor={some author},pdftitle={eye-catching title}}
\ifCLASSINFOpdf
\else
\fi
\usepackage{fixltx2e}

\usepackage{stfloats}
\begin{document}
%
\title{Multiple Infrared Small Targets Detection based on Hierarchical Maximal Entropy Random Walk}
%
%
%

\author{Chaoqun~Xia, ~Xiaorun~Li, ~Liaoying Zhao
\thanks{This work is supported by the National Nature Science Foundation of China (No.61671408), Shanghai Aerospace Science and Technology Innovation Fund (No.SAST2015033) and the Joint Fund of the Ministry of Education of China (No.6141A02022314). (\emph{Corresponding author: Xiaorun Li.})}
\thanks{C. Xia and X. Li are with the College of Electrical Engineering, Zhejiang University, Hangzhou 310027, China (e-mail: auto\_xia@foxmail.com; lxrly@zju.edu.cn)}
\thanks{L. Zhao is with the Institute of Computer Application Technology,
Hangzhou Dianzi University Hangzhou, 310018, China (e-mail:
zhaoly@hdu.edu.cn).}}

%
%

\markboth{}%
{}
%



\maketitle

\begin{abstract}
The technique of detecting multiple dim and small targets with low signal-to-clutter ratios (SCR) is very important for infrared search and tracking systems. In this paper, we establish a detection method derived from maximal entropy random walk (MERW) to robustly detect multiple small targets. 	Initially, we introduce the primal MERW and analyze the feasibility of applying it to small target detection. However, the original weight matrix of the MERW is sensitive to interferences. Therefore, a specific weight matrix is designed for the MERW in principle of enhancing characteristics of small targets and suppressing strong clutters. Moreover, the primal MERW has a critical limitation of strong bias to the most salient small target. To achieve multiple small targets detection, we develop a hierarchical version of the MERW method. Based on the hierarchical MERW (HMERW), we propose a small target detection method as follows. First, filtering technique is used to smooth the infrared image. Second, an output map is obtained by importing the filtered image into the HMERW. Then, a coefficient map is constructed to fuse the stationary dirtribution map of the HMERW. Finally, an adaptive threshold is used to segment multiple small targets from the fusion map. Extensive experiments on practical data sets demonstrate that the proposed method is superior to the state-of-the-art methods in terms of target enhancement, background suppression and multiple small targets detection.
\end{abstract}


%
\IEEEpeerreviewmaketitle

\section{Introduction}
\label{sec:introduction}
%
%
%
%
In the fields of remote sensing, space surveillance and antiaircraft warning systems,  Infrared Search and Track (IRST) system has been widely used and drawn much attention \cite{MORADI2018387}. Infrared small target detection is one of the key techniques in the IRST system. The task of infrared small target detection is to detect moving objects at a long distance. Because of the long-distance imaging, the targets photographed by infrared sensors occupy only a few pixels in the infrared images. This results in the scarcity of texture and spatial characteristics for target detection. In addition, infrared sensors often generate heavy noise during the long-distance imaging \cite{hunt1974infrared}. This inherent sensor noise may also produce pixel-sized noise with high brightness (PNHB) \cite{8660588}. Moreover, the moving targets are usually buried in intricate backgrounds and varying clutters. For example, some practical detection scenes include aircrafts flying across heavy clouds, cars running on complex roads and boats sailing on the sea. Under these complicated scenes, the signal-to-clutter ratios of infrared images are quite low. All the above adverse conditions make small target detection a difficult and challenging problem.

To solve the problem, researchers have made great efforts and achieved some promising progress. So far, numerous small target detection methods have been proposed. The existed detection methods can be generally divided into two categories, sequential detection methods \cite{DENG2018122,GAO2018463,8698272,8374078,8867952} and single-frame detection methods. In contrary to single-frame detection methods, sequential detection methods employ temporal cues to extract moving targets from static backgrounds. However, these methods may not work if the targets move quickly through varying clutters. Besides, the sequential detection methods generally cost more computation time than single-frame detection methods because they process multiple frames for detection \cite{8705367}. The above drawbacks make the sequential detection methods unsuitable for the IRST systems deployed on the space surveillance and antiaircraft warning systems. In this case, more attention has been paid to the single-frame detection methods.

In the past decades, numerous single-frame detection methods have been invented. For convenience of illustration, the existed single-frame detection methods can be organized into four groups according to the techniques they use, including filtering skill \cite{deshpande1999max,KIM2012393,7867372,6010000,Li2014,BAI20102145,Deng2018}, human vision system \cite{6479296,rs11010014,WEI2016216,NIE2018186,8289318,7460907,8245877,8723140,LI2018113}, sparse representation theory \cite{6595533,7365444,DAI2017182,HE201598,rs10111821,LI2017238} and statistical learning theory \cite{4770453,rs10122004,10.1117/12.2520250,GAO2019206}. 

Two of the earliest filters used for small target detection are max-mean and max-median filters \cite{deshpande1999max}, proposed by S. D. Deshpande et al. in 1999. More recently, S. Kim et al. \cite{KIM2012393} proposed a Laplacian of Gaussian filter to optimize signal-to-clutter ratio in heterogeneous background. Variants of top-hat filter \cite{BAI20102145,Deng2018} have also been applied to suppress clutters for small target detection. In addition to the spatial filters, some frequency-domain filters based on Fourier transform \cite{7867372} and wavelet transform \cite{6010000} were invented to eliminate background distributed in the low frequency domain. Generally, the filter-based small target detection methods make the same assumption that the backgrounds and clutters are spatially heterogeneous or distributed in the low frequency domain while the targets are inhomogeneous or distributed in the high frequency domain. However, this assumption is not true for small targets submerged in heavy clutters and noise.

Human vision system (HVS) manages to encode the contrast feature in the streams of visual system and has found wide application in computer vision tasks \cite{6479296}. As small targets usually exhibit local contrast feature, some researchers attempted to encode the feature by designing specific local descriptors. Inspired by the HVS and derived kernel, Kim et al. \cite{6479296} proposed a descriptor called local contrast measure (LCM) to enhance the local contrast feature of small targets. Based on this work, several variants of the LCM such as multi-scale relative local contrast measure (MRLCM) \cite{8289318}, weighted local contrast measure (WLCM) \cite{7460907} and homogeneity-weighted local contrast measure (HWLCM)  have been presented. Likewise, other advanced local descriptors based on derivative entropy \cite{8245877}, local energy factor (LEF) \cite{8723140} and local steering kernel (LSK) reconstruction \cite{LI2018113} were proposed and applied to enhance small targets. Although HVS-based small target detection methods perform well in target enhancement, they struggle to suppress strong interferences and pixel-size noise of high brightness (PNHB) that reveal similar local contrast feature.

Sparse representation and component analysis theories were prevalently used for image processing and target detection in the past years \cite{5711635}. By assuming that the small target appears in the sparse component while the background belongs to the low-rank component, some researches \cite{7365444,HE201598} intended to search the target in a sparse matrix recovered from the input infrared image. Gao et al. \cite{6595533} further studied the robust principle component analysis (RPCA) and proposed a small target detection scheme based on it. Likewise, other small target detection methods based on, e.g., partial sum minimization of singular values \cite{DAI2017182}, non-convex rank approximation minimization joint $\ell_{2,1}$ norm \cite{rs10111821} and singular value decomposition \cite{LI2017238}, were also proposed. Compared with the filter-based and HVS-based methods, the sparse representation-based methods can eliminate more clutters and better enhance the small targets. However, they are quite sensitive to PNHB because PHNB is generally found in the sparse component.

Motivated by the advance of applying statistical learning theory to computer vision tasks, many learning-based small target detection methods were developed. Wang et al. \cite{LI2017238} considered the detection task as a binary classification problem and used least square support vector machine to distinguish between targets and backgrounds. Moreover, random walk model was first applied to detect small targets by Xia et al. \cite{rs10122004}, and then Qin et al. \cite{8705367} improved the method and developed a local version of random walk. In addition, neural networks \cite{GAO2019206} and deep learning theory \cite{10.1117/12.2520250} were also employed for small target detection. These methods were data-driven, however,  in most cases, prior data is unreachable for IRST systems.

\begin{figure*} [!t]
\centering 
\includegraphics[width=\textwidth]{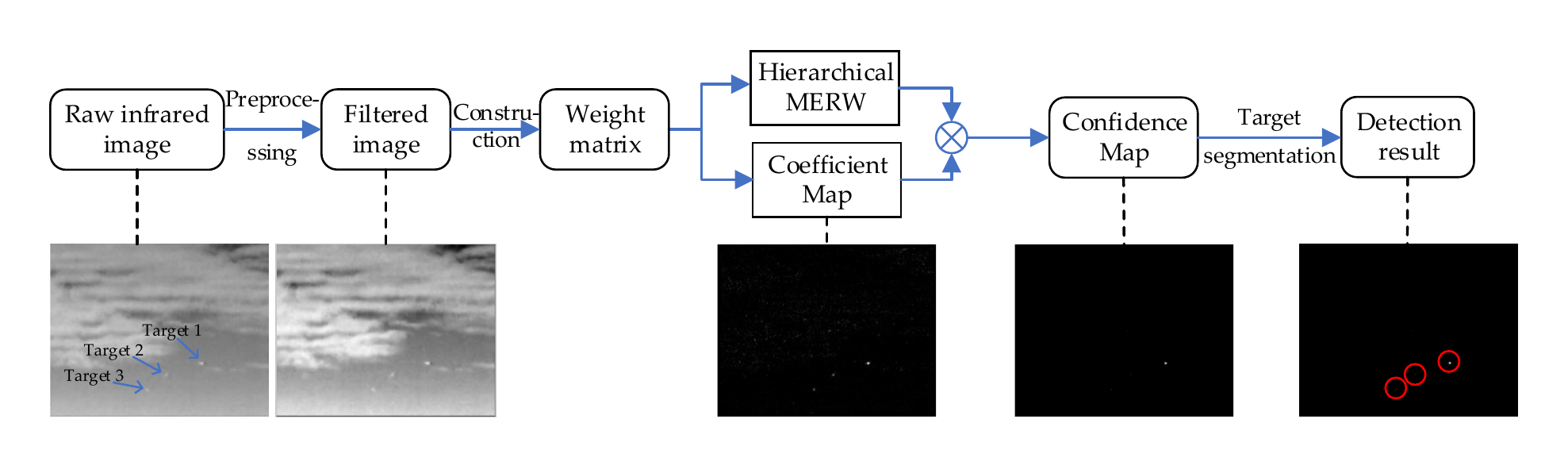}
\caption{The framework of the proposed small target detection method. (Pseudo-color images are transformed from gray-scale images.)}
\label{fig:framework}
\end{figure*}

Although great progress has been made in the past years, there still remains an open issue of designing a small target detection method robust to various detection scenes. Additionally, few studies investigated detection scenes with multiple small targets. In some applications, for example, in the drone groups detection scenario, there is more than one target of interest to be detected. In such scenarios, the contrast between different small targets and background is different, some are distinct while others are dim. Fig. \ref{fig:framework} shows a practical infrared image with three small targets, of which Target 2 and Target 3 are much dimmer than Target 1. The existed small target detection methods generally fail to detect these two dim targets under such intricate clutters, as verified in Section \ref{sec:experiment}.

To fix this issue, we design a small target detection method that is robust to diverse scenes, including multiple small targets detection. As stated in the previous studies \cite{rs10122004,8705367}, small targets reveal specific characteristics, including contrast consistency, regional compactness and global uniqueness. To be specific, the contrast consistency characteristic indicates that pixels of a small target have a signature of omni-directional intensity discontinuity compared with their neighboring pixels. The regional compactness is that the intensities of the target pixels are similar to each other. The global uniqueness means that the distribution of small targets in the image is random and relatively rare. Accordingly, image regions that reveal these three characteristics can be considered as small targets.

With the above considerations in mind, our detection method is designed on the basis of the maximal entropy random walk (MERW). It has been proved in \cite{rs10122004} that random walk was effective for measuring global uniqueness. Compared with the generic random walk, the MERW is a revised version in principle of maximizing the Shannon entropy of the random walk process. The framework of the proposed detection method is shown in Fig. \ref{fig:framework}. First, mean filtering technique is used to smooth the noisy image. Second, the filtered image is transported to the MERW to build a confidence map. Instead of using the primal MERW directly, we develop a hierarchical version for multiple small targets detection. Besides, a specific weight matrix is designed for the HMERW. The intension of the designed weight matrix is to enhance the contrast consistency and regional compactness characteristics of small targets. Then, the stationary distribution map of the HMERW is further fused with a coefficient map, which is constructed based on the designed weight matrix. Finally, the targets can be readily segmented from the background using an adaptive threshold.

The advances of this research can be summarized as follows. First, our research focus on resolving a specific task of multiple small targets detection, which is a common scenario for IRST systems, and few studies achieve satisfactory detection performance in this task. Second, to authors' knowledges, this is the first research that analyzes the limitation of the primal MERW model for practical applications. Moreover, we evaluate the adverse effects of this limitation for practical applications and provide a theoretical interpretation. Third, we develop a HMERW to solve the limitation of the MERW based on a graph decomposition theory, which manages to analyze a graph in different subspaces, and a specifically designed weight matrix, which describes local characteristics of small targets. It is worthwhile to note that the graph decomposition theory is applicable to other graph-based models. Finally, we design a multiple small targets detection method based on the HMERW and fusion operation. Extensive experiments on several practical data sets demonstrate superiority of the proposed method in terms of target enhancement, background suppression and multiple small targets detection.

The remainder of this paper is organized as follows. Section \ref{sec:overview-MERW} summarizes the most related work of MERW. Section \ref{sec:motivation} analyzes motivation and limitation of MERW for small target detection. The HMERW is presented in Section \ref{sec:HMERW}. Section \ref{sec:methodology} introduces the proposed small target detection method in detail. Section \ref{sec:novelty} is particularly included to highlight contributions and novelties of HMERW. Extensive experiments and discussion are given in Section \ref{sec:experiment}. Conclusions are drawn in Section \ref{sec:conclusion}.

\section{Related work}\label{sec:overview-MERW}
In this section, we briefly review the primal MERW for image processing. Given a single-band image $\bf{I}\in\mathbb{R}^{\emph{m}\times\emph{n}}$, build an undirected and weighted graph for $\bf{I}_{\emph{m}\times\emph{n}}$ as $\mathcal{G=(V,E)}$, where $\mathcal{V}=\{v_1,\dots,v_N\}$ and $\mathcal{E}=\{e_{ij}|v_i,v_j\in\mathcal{V}\}$ are sets of $N=m\times n$ nodes and their edges, respectively. The weight matrix associated to $\mathcal{E}$ is denoted by $\bf{W}\in\mathbb{R}^{\emph{N}\times\emph{N}}$ such that $\textbf{W}_{ij}=\textbf{W}_{ji}$, due to the undirected edges. Random walks of length $t$ on a graph can be viewed as a Markov process with random state variables $\{X_i|i=1,\dots,t,X_i\in\mathcal{V}\}$. It has been proven in \cite{tierney1996introduction} that a Markov process on an undirected graph is irreducible and aperiodic. Suppose the probability that a walker jumps from node $v_i$ to node $v_j$ is $\textbf{P}_{ij}$ besides satisfying $\sum_{j}\textbf{P}_{ij}=1$ for each node $v_i$. The major interest of a random walk model is to predict the probability of finding the walker at node $v_i$ after taking infinite steps ($t\to\infty$). According to the Ergodic theory, if a Markov chain is irreducible and not periodic, there exists a unique stationary distribution $\boldsymbol{\pi}\in\mathbb{R}^{1\times N}$ such that 
\begin{eqnarray}\label{eq:stationary condition}
\boldsymbol{\pi}=\boldsymbol{\pi}\textbf{P},\sum_{v_i\in\mathcal{V}}\boldsymbol{\pi}_i=1.
\end{eqnarray}

It has been stated in \cite{PhysRevLett.102.160602} that attaining uniform distribution along paths of any length leads to maximal entropy for the random walk process. By viewing the weight $\textbf{W}_{ij}$ as the number of pathways connecting node $v_i$ and node $v_j$, there are $(\textbf{W}^{t})_{ij}$ pathways from node $v_i$ to node $v_j$ after taking $t$ steps, where $\textbf{W}^{t}$ denotes the $t$-th power of the matrix $\textbf{W}$. Accordingly, the MERW defines the transition matrix as
\begin{eqnarray}\label{eq:MERWprob}
\textbf{P}_{ij}^\text{MERW}=\lim_{t\to\infty}\frac{\textbf{W}_{ij}\sum_{v_k\in\mathcal{V}}(\textbf{W}^{t-1})_{jk}}{\sum_{v_l\in\mathcal{V}}\textbf{W}_{il}\sum_{v_k\in\mathcal{V}}(\textbf{W}^{t-1})_{lk}}.
\end{eqnarray}
Following the Perron-Frobenius theorem, $\textbf{W}^{t}$ with $t\to\infty$ can be asymptotically approximated by 
\begin{eqnarray}\label{eq:initialW}
\textbf{W}^t\approx\lambda_1^t\boldsymbol{\psi}^{(1)}{(\boldsymbol{\psi}^{(1)})}^\text{T},
\end{eqnarray}
where $\lambda_1$ is the largest eigenvalue of $\textbf{W}$, and $\boldsymbol{\psi}^{(1)}\in\mathbb{R}^{N\times1} \text{ satisfies } {(\boldsymbol{\psi}^{(1)})}^\text{T}\boldsymbol{\psi}^{(1)}=1$ and $\textbf{W}\boldsymbol{\psi}^{(1)}=\lambda_1\boldsymbol{\psi}^{(1)}$. By substituting \eqref{eq:initialW} into \eqref{eq:MERWprob}, the transition matrix of the MERW reads
\begin{eqnarray}\label{eq:MERWP}
\textbf{P}^{\text{MERW}}_{ij}=\frac{\textbf{W}_{ij}\boldsymbol{\psi}^{(1)}_j}{\lambda_1\boldsymbol{\psi}^{(1)}_i},
\end{eqnarray}
where $\boldsymbol{\psi}_i^{(1)}$ is the i-th element of vector $\boldsymbol{\psi}^{(1)}$. Then, the stationary distribution of the MERW is
\begin{eqnarray}\label{eq:MERW stationary}
\boldsymbol{\pi}^{\text{MERW}}_i={(\boldsymbol{\psi}^{(1)}_i)}^2,
\end{eqnarray}
which can be easily derived by combining \eqref{eq:stationary condition} and \eqref{eq:MERWP}. 


In recent years, the MERW has been successfully applied to salient object detection \cite{6678551,7574714}.
 The success of this application is due to the inherent properties of the MERW. First, the walker is biased to visit nodes with larger degrees (note the degree of a node is the number of pathways connecting it and its neighboring nodes). By taking image regions (or called super-pixels) as nodes and their dissimilarities as weights, it is more possible to place the walker at salient nodes after reaching equilibrium, as illustrated in \cite{7574714}. In other words, a salient node will have a large value in the stationary distribution. Second, the walker of the MERW takes into consideration the global knowledge when deciding the pathway. The awareness of global knowledge can be perceived from the definition of transition probability in \eqref{eq:MERWprob}. More specifically, suppose $\sum_{u_1,\dots,u_t\in\mathcal{V}}\text{Path}(v_i,u_1,\dots,u_{t-1},u_t)$ denotes the number of all possible $t$-length pathways starting from node $v_i$, the MERW defines the probability of moving from node $v_i$ to node $v_j$ as a ratio of $\sum_{u_2,\dots,u_t}\text{Path}(v_i,v_j,u_2,\dots,u_{t-1},u_t)$ to $\sum_{u_1,\dots,u_t\in\mathcal{V}}\text{Path}(v_i,u_1,\dots,u_{t-1},u_t)$, with $t$ approaching to infinity. This way, global structure of the graph is utilized in the walking procedure.

\begin{figure}[htbp]
\centering 
\subfloat[A synthetic data set.]{
\label{fig:synthesis-data}
\includegraphics[width=\columnwidth]{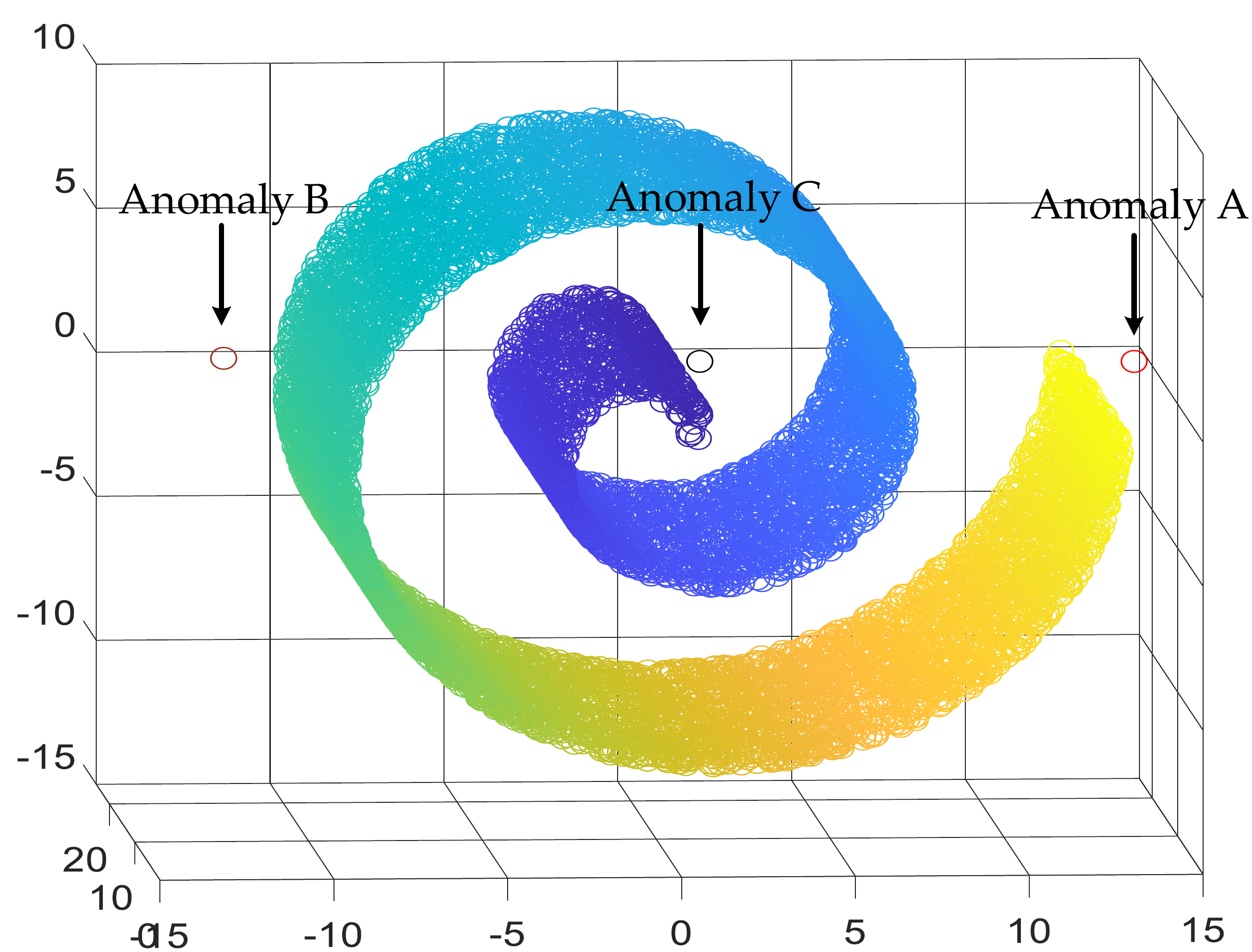}}\hfill
\subfloat[Stationary distribution of MERW.]{
\label{fig:synthesis-k-1}
\includegraphics[width=\columnwidth]{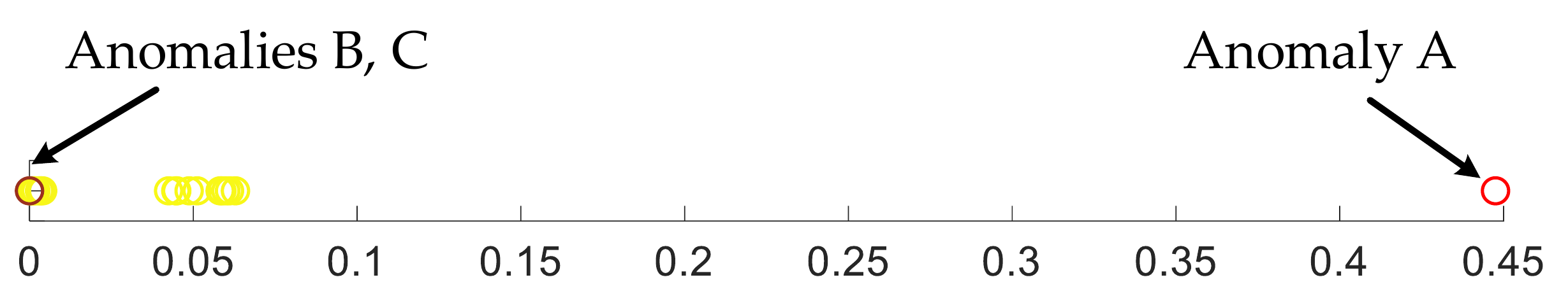}}\hfill
\subfloat[Stationary distribution of \eqref{eq:HMERW stationary} with $K=2$.]{
\label{fig:synthesis-k-2}
\includegraphics[width=\columnwidth]{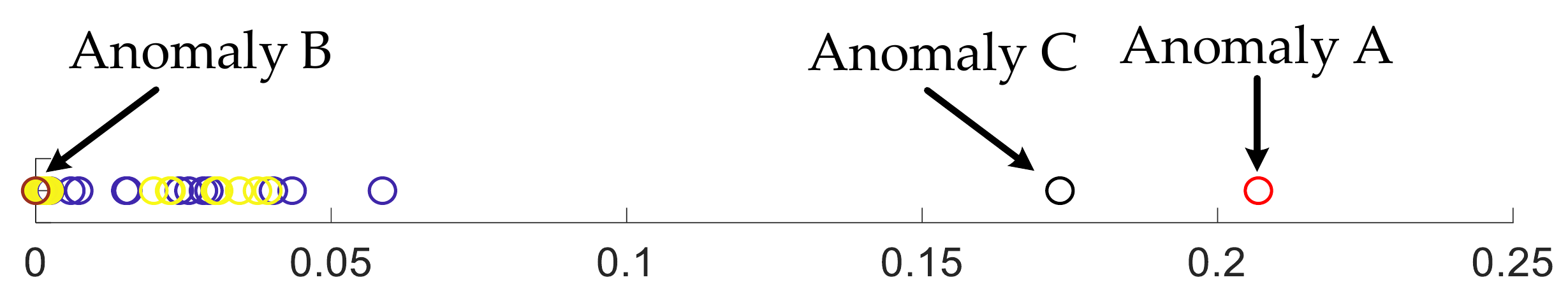}}\hfill
\subfloat[Stationary distribution of \eqref{eq:HMERW stationary} with $K=3$.]{
\label{fig:synthesis-k-3}
\includegraphics[width=\columnwidth]{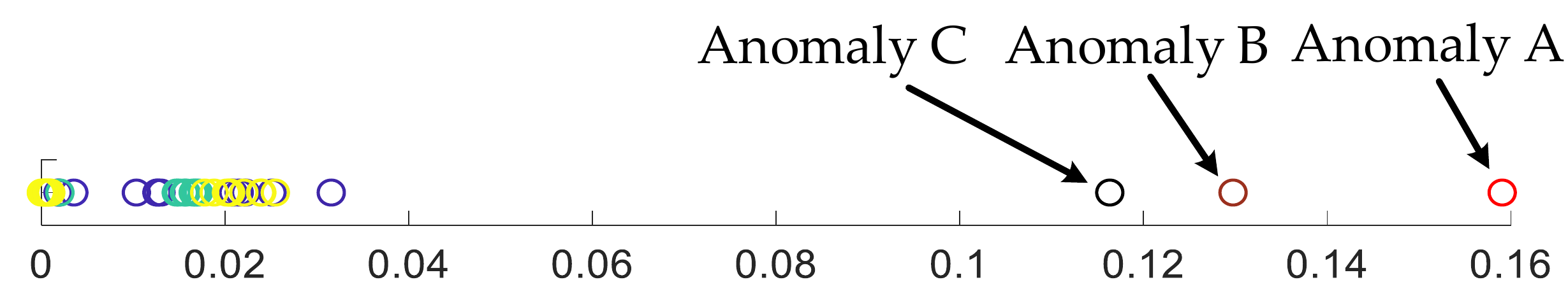}}\hfill
\subfloat[Stationary distribution of \eqref{eq:HMERW stationary} with $K=50$.]{
\label{fig:synthesis-k-50}
\includegraphics[width=\columnwidth]{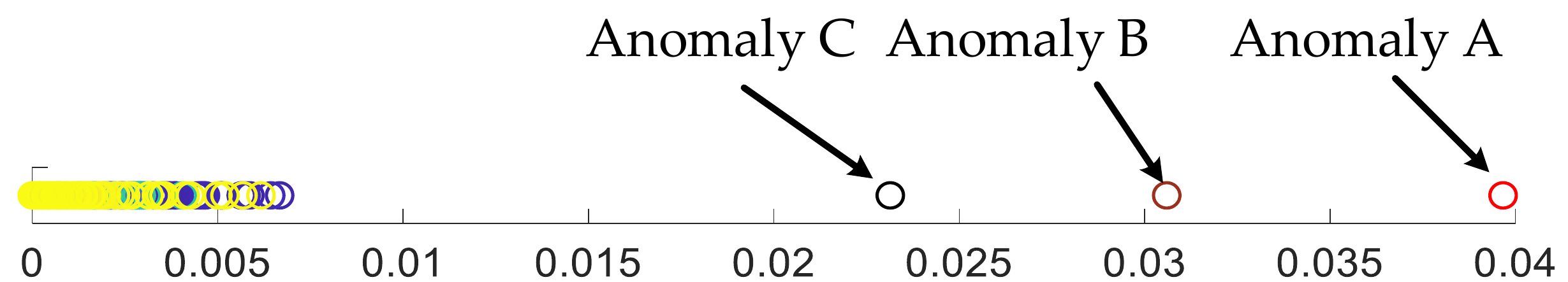}}
\caption{Simulation results of maximal entropy random walk (MERW) for expressing the global uniqueness of anomalies and our hierarchical version for separating multiple anomalies from homogeneous nodes. (a) A synthetic manifold data set with 10000 nodes and 3 anomalies. (b) One-dimensional embedding for the synthetic data set obtained by \eqref{eq:MERW stationary}. (c)-(e) are stationary distributions of HMERW corresponding to \eqref{eq:HMERW stationary} with $K=2,3$ and 50, respectively.}
\label{fig:synthesis}
\end{figure}

\section{Motivation and limitation of MERW for small target detection}\label{sec:motivation}
As stated in Section. \ref{sec:overview-MERW}, by maximizing Shannon entropy during the Markov process, the MERW model owns the ability to exploit global correlation of nodes and quantify it in its stationary distribution. In other words, the more anomalous the node is, the larger the stationary probability $\boldsymbol{\pi}^\text{MERW}_i$ is. Generally, a small target can be viewed as an anomalous target due to one of its specific characteristic, global uniqueness. Specifically, anomalies are relatively distinct and show weak correlation with the pixels of background. Likewise, small targets reveal dissimilarity to the homogeneous background and are scarce in the image. Due to this analogy, we reasonably believe that the MERW model can enhance the global characteristic of small targets if it can separate anomalies away from background in the stationary distribution. To verify it, we perform a simulation experiment on a synthetic data set.

\paragraph{Synthetic data set and settings}  As shown in Fig. \ref{fig:synthesis-data}, the synthetic data set comprises a swiss roll with 10000 random nodes (represented as background) and 3 anomalies, denoted by A, B, C, in the vicinity. Note that Euclidean distance between anomalies and their corresponding nearest nodes is $d_A,d_B,d_C$ satisfying $d_A>d_B=d_C$. To begin with, we build an undirected and weighted graph for the synthetic data. In the graph, each node connects to its 10 nearest neighbors, and the weight is quantified by Euclidean distance. Then, we import this graph into the primal MERW model and educe the stationary distribution according to \eqref{eq:MERW stationary}. 

\paragraph{Results and analysis}
The output of the MERW model is displayed in Fig. \ref{fig:synthesis-k-1}. Several considerable phenomena are observed as follows.
\begin{enumerate}
\item[1)] Anomaly A is pushed away from other nodes.
\item[2)] 10 nearest nodes of Anomaly A (yellow-tinted nodes spreading near the value of 0.05 in Fig. \ref{fig:synthesis-k-1}) depart from the other nodes on the swiss roll.
\item[3)] Anomalies B and C blend with the background nodes, excluding Anomaly A and its 10 nearest nodes.
\end{enumerate}

The first phenomenon provides evidence for the validity of the MERW model in enhancing the global uniqueness of anomalies, that is, the MERW model can enhance the most significant anomalies and suppress other nodes. The second and third phenomena result from the limitation of the primal MERW model.
The derivation of the stationary distribution of \eqref{eq:MERW stationary} relies on a crucial assumption, that is, it takes infinite steps ($t\text{ of }\textbf{W}^t$ approaches +$\infty$) for a walker to reach equilibrium on an irregular graph. Based on this assumption, $\textbf{W}^t$ can be approximated by \eqref{eq:initialW}, which leads to a simplified transition matrix of \eqref{eq:MERWP} and a stationary distribution of \eqref{eq:MERW stationary}. However, this assumption generally fails if a graph comprises of numerable nodes, which is common in practical scenarios. In this case, a walker only takes countable steps to reach equilibrium. Therefore, the reconstruction error of using the principle component to approximate $\textbf{W}^t$ appears to be considerable, making the stationary distribution of \eqref{eq:MERW stationary} less convincing. 

As mentioned before, the weights between Anomaly A (the most distinct anomaly) and its 10 nearest nodes are much larger than those weights between Anomalies B, C and their neighbors. The primal MERW approximates the weight matrix with the principle component, which only magnifies the dominant information of $\textbf{W}$ (i.e., information of weights connecting to Anomaly A, we will explain this in the next section). Then, the dominant information is further intensified and the remaining information (e.g., information of weights between inner nodes and Anomalies B, C) is suppressed by $\textbf{W}^t$ of the transition matrix in \eqref{eq:MERWprob}. Accordingly, due to the effect of the intensified information connecting to Anomaly A, the 10 nearest nodes of Anomaly A attain larger stationary probability than the remaining background nodes and Anomalies B, C. Besides, information with regard to Anomalies B, C and the background nodes (except for Anomaly A and its 10 nearest nodes) are suppressed, leading to their small stationary probabilities in Fig. \ref{fig:synthesis-k-1}. 

\paragraph{Conclusion}Based on the simulation experiments, we can draw the following conclusions. First, the primal MERW model is applicable to enhance global characteristic of small targets, according to phenomenon 1). Second, the primal MERW has its major limitation of strong bias to nodes with dominant information. Derived from this limitation, the primal MERW will discard some valuable information of important nodes and fail to discriminate sub-anomalous nodes from background nodes. Accordingly, it remains further exploitation of the MERW model for enhancing multiple small targets.

\begin{figure} [!h]
\centering 
\includegraphics[width=0.8\columnwidth]{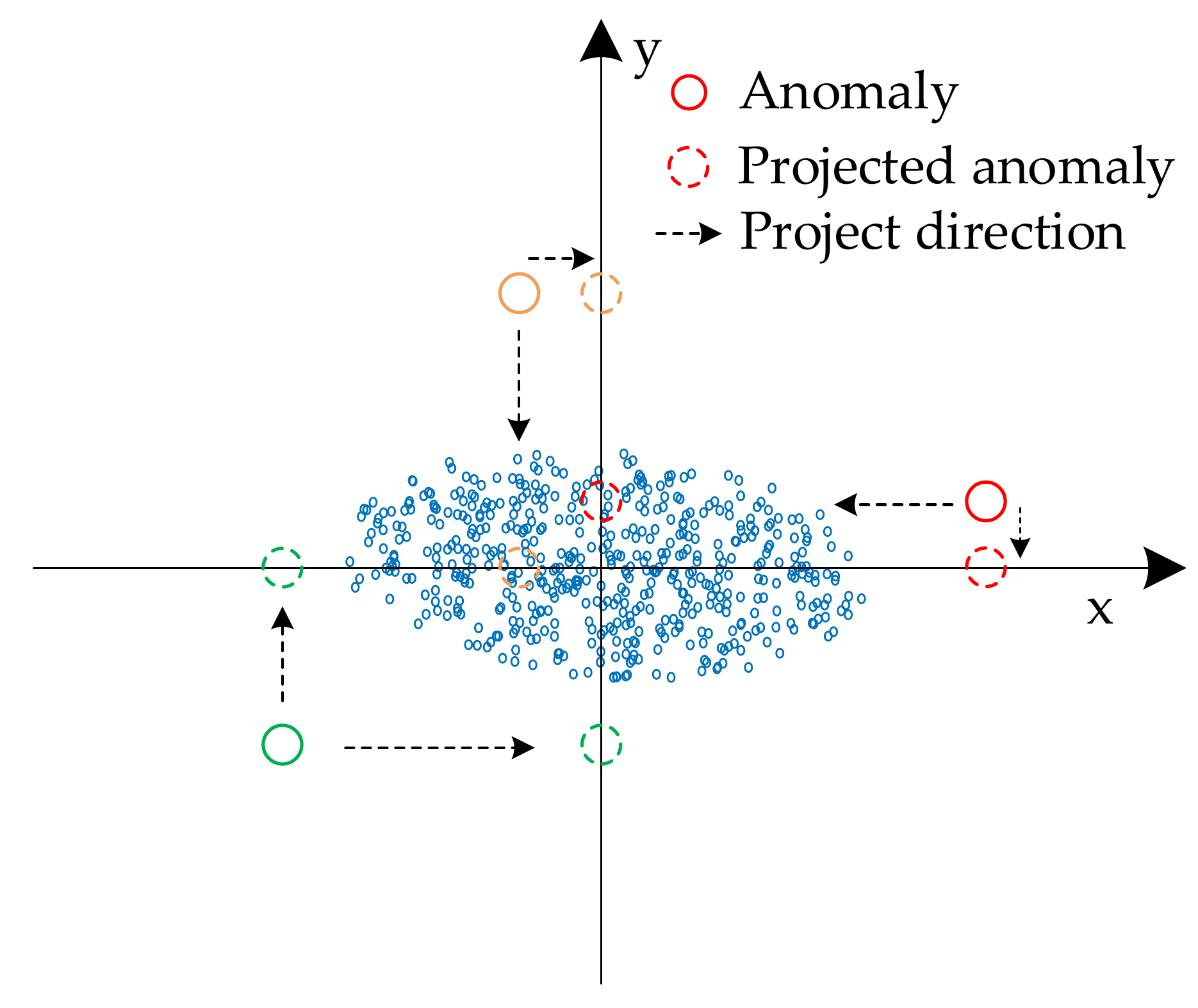}
\caption{Schematic diagram of anomaly projection.}
\label{fig:PCA}
\end{figure}
\section{HMERW}\label{sec:HMERW}
In this section, we look into the theoretical limitation of the primal MERW for practical applications, and develop a hierarchical version of MERW that makes up the limitation, which has never been explored before.

As analyzed in Section \ref{sec:motivation}, the limitation of the primal MERW derives from the insufficient reconstruction of weight matrix in \eqref{eq:initialW}. 
In essence, \eqref{eq:initialW} manages to project the weight matrix along a specific direction, determined by the principle eigenvector, in the principle of maximizing variance of the projected data. For this view, the limitation of strong bias to nodes with dominant information can be theoretically explained. 

Taking Fig. \ref{fig:PCA} as an illustrative example, there are three anomalies and a group of nodes randomly generated within an ellipse in a two-dimensional coordinate space. Following the maximal variance principle, we will priorly choose to project the nodes onto the x-axis. This way, the red anomaly moves far away from the other nodes, in contrary to the yellow anomaly being submerged in the homogeneous nodes. This phenomenon is consistent with the synthetic experiments in Fig. \ref{fig:synthesis-k-1}. However, after we project the nodes onto the y-axis, which is orthogonal to x-axis, the yellow anomaly is separated from the node cluster while the red anomaly is submerged. This implies that the primal MERW shows bias to nodes that contribute significantly to projection variance. 
In the case of Fig. \ref{fig:PCA}, the primal MERW tends to emphasize the red anomaly and conceal the information of the yellow anomaly. 
Moreover, the green anomaly in Fig. \ref{fig:PCA} can be separated from the node cluster in spite of the projection directions. 
This suggests that the information of an anomaly will be enhanced if its deviation direction from the node cluster is consistent with the projection direction. Otherwise, the information of an anomaly will be suppressed. For instance, the direction in which the red anomaly deviates from the blue nodes cluster is the x-axis direction. If the red circle is projected to the x-axis direction, the anomaly information will be enhanced; else if it is projected to the orthogonal space of the x-axis, that is, the y-axis, the anomaly information will be suppressed.
Another promising finding from Fig. \ref{fig:PCA} is that paths connecting two nodes are orthogonal in different projection subspaces. In this manner, random walk process on nodes projected in orthogonal subspaces is independent of each other. In what follows, we develop a graph decomposition theory based on the aforementioned observations.

\newtheorem*{prop}{Graph Decomposition}
\begin{prop}
Given a graph $\mathcal{G}=(\mathcal{V},\mathcal{E},\textbf{W})$  spanned on a space $\mathcal{S}$, a graph $\mathcal{G}^{(k)}=(\mathcal{V}^{(k)},\mathcal{E}^{(k)},\textbf{W}^{(k)})$ obtained by projecting $\mathcal{G}$ onto a subspace $\mathcal{S}_k=\text{Span}\{\textbf{p}^{(k)}\}$ is called the k-level sub-graph of $\mathcal{G}$ if it satisfies
\begin{eqnarray}\label{eq:optimal-direction}
\min_{\textbf{W}^{(k)}}||\textbf{W}-\sum_{i=1}^{k-1}\textbf{W}^{(i)}-\textbf{W}^{(k)}||^2_2,\\
s.t.\ \textbf{W}^{(k)}=\textbf{p}^{(k)}{(\textbf{p}^{(k)})}^\text{T}, (\textbf{p}^{(k)})^\text{T}\textbf{p}^{(i)}=\textbf{0},
\end{eqnarray}
for $i=1,...,k-1$ and $k=1,...,\text{rank}(\textbf{W})$.
The optimal direction is given by $\textbf{p}^{(k)}=\sqrt{\lambda_k}\boldsymbol{\psi}^{(k)}$, where $(\lambda_k, \boldsymbol{\psi}^{(k)})$ is the k-th eigenpair of $\textbf{W}$.
\end{prop}

It can be perceived from the objective function \eqref{eq:optimal-direction} that the criterion of $k$-level graph projection is to minimize the information loss of the weight matrix, and the optimal direction can be easily derived according to Lagrange multiplier. Based on the graph decomposition theory, a hierarchical version of maximal entropy random walk (HMERW) model is developed as follows.
\begin{enumerate}
\item[1)] Decompose the primal graph $\mathcal{G}$ into $K$ orthogonal sub-graphs, and weight matrix of the $k$-level sub-graph $\mathcal{G}^{(k)}$ is determined by $\textbf{W}^{(k)}=\lambda_k\boldsymbol{\psi}^{(k)}(\boldsymbol{\psi}^{(k)})^\text{T}$,  where $(\lambda_k, \boldsymbol{\psi}^{(k)})$ is the k-th eigenpair of $\textbf{W}$.
\item[2)] Perform maximal entropy random walk on each sub-graph and obtain the stationary distribution by
\begin{eqnarray}\label{eq:part-pi}
\boldsymbol{\pi}^{(k)}_i=\lambda_k{(\boldsymbol{\psi}^{(k)}_i)}^2.
\end{eqnarray}
\item[3)] Reconstruct the stationary distribution of the primal graph as

\begin{eqnarray}\label{eq:HMERW stationary}
\boldsymbol{\pi}^\text{HMERW}_i=\frac{\sum_{k=1}^K\lambda_k(\boldsymbol{\psi}^{(k)}_i)^2}{\sum_{j=1}^N\sum_{k=1}^{K}\lambda_k(\boldsymbol{\psi}^{(k)}_j)^2}.
\end{eqnarray}
Note that $\sum_{i=1}^N\boldsymbol{\pi}^\text{HMERW}_i=1$ and $\boldsymbol{\pi}_i^\text{HMERW}>=0$ because of $\lambda_k>=0$ for the semidefinite matrix \textbf{W}. 
\end{enumerate}

To verify the validity of the proposed theory of graph decomposition, we perform our HMERW on the synthetic dataset of Fig. \ref{fig:synthesis-data}. As shown in Fig. \ref{fig:synthesis-k-2}, \ref{fig:synthesis-k-3} and \ref{fig:synthesis-k-50},  we display the stationary distributions of HMERW by decomposing the primal graph into 2, 3 and 50 orthogonal sub-graphs, respectively. As observed in Fig. \ref{fig:synthesis-k-2}, apart from the prominent Anomaly A, Anomaly C is pulled away from the node cluster after we take an additional sub-graph into consideration for MERW. Likewise, after importing information of the third sub-graph, Anomaly B is separated from the cluster, as shown in Fig. \ref{fig:synthesis-k-3}. It is noteworthy that the inner nodes (blue-tinted nodes on the swiss roll) gradually move apart from the outer nodes (yellow-tinted nodes) and exhibit larger stationary probability when decomposing the primal graph into more orthogonal sub-graphs, as shown in Fig. \ref{fig:synthesis-k-50}. It is found that the interior of the swiss roll has larger curvature than the exterior, thus the Euclidean distance between the inner nodes tends to be larger than that of the outer nodes. Therefore, the inner nodes are supposed to show higher stationary probability than the outer node since a walker has more path ways to arrive at the inner nodes. For this view, the developed HMERW succeed in exploring such a subtle feature by involving enough information of sub-graphs. 

\section{Multiple Small Targets Detection Method based on HMERW} \label{sec:methodology}  

In this section, the proposed small target detection method based on the HMERW is introduced in detail. First, the original infrared image is preprocessed. Next, we construct an HMERW model for the filtered image and further improve the HMERW by designing a specific weight matrix in principle of enhance characteristics of a small target. After that, a coefficient map is built based on the designed weight matrix and used to fuse the stationary distribution map of the HMERW. Then, an adaptive segmentation method is used to extract small targets from the fusion map. Finally, a detailed algorithm of the proposed detection scheme is given.

\subsection{Preprocessing}\label{sec:Preprocessing}
Usually, remote sensing infrared images have strong noise. Thus, it is difficult to directly apply a random walk model to process raw intensities of the noisy images. In this case, a $2\times 2$ mean filter is used to smooth the raw images. The mean filter is applied for two reasons, making the image show better consistency and suppressing the intensities of the small-scale (less than $2\times2$ pixels) PNHBs. The mean filtering technique will make the infrared images more suitable for our random walk model. For simplicity of illustration, the filtered infrared image is referred as \textbf{I} hereinafter.

\subsection{Weight matrix for the HMERW} \label{sec:MMERW4}

\begin{figure}[htbp]
\centering 
\subfloat[]{
\label{fig:examples-a}
\includegraphics[width=0.5\columnwidth]{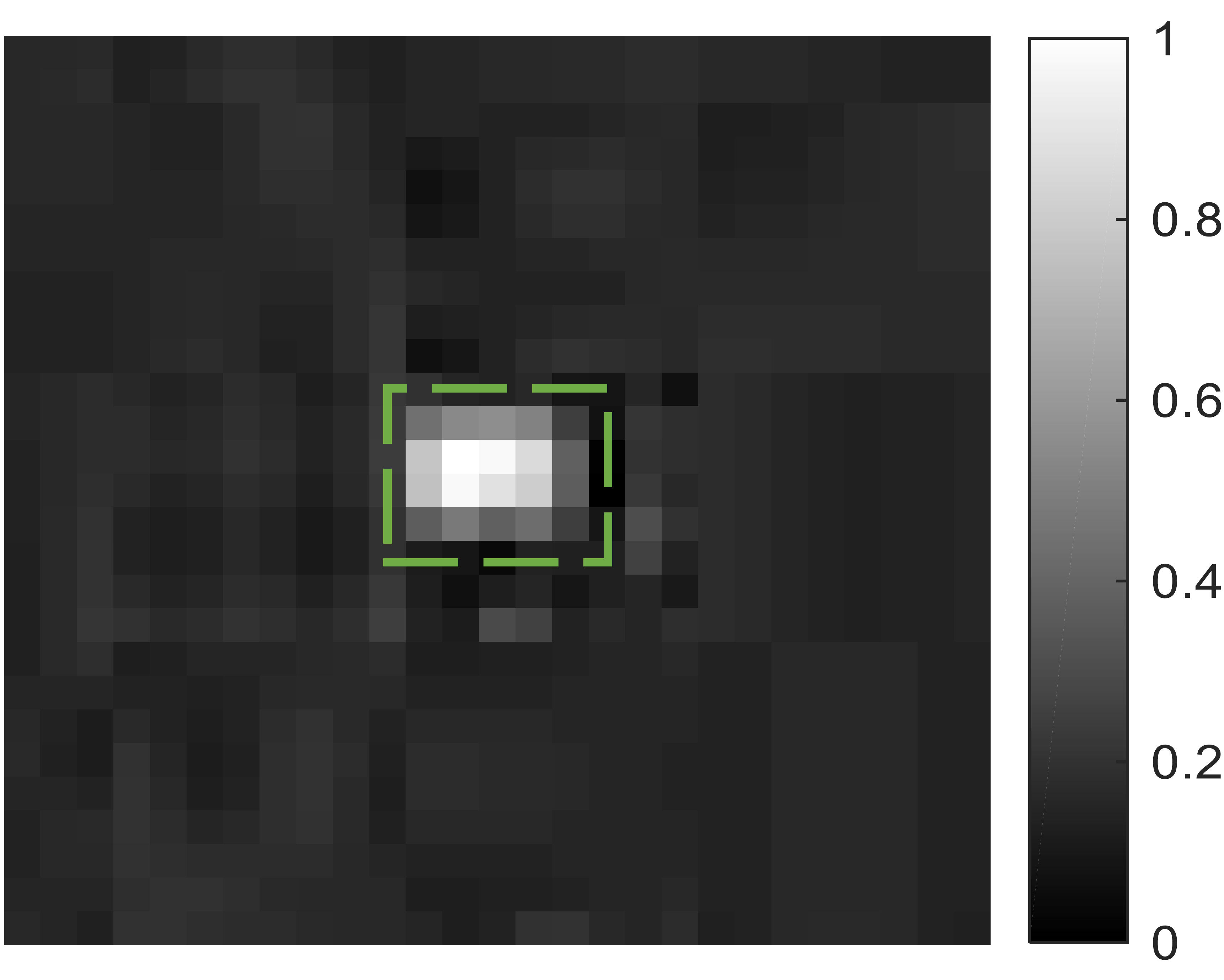}}
\subfloat[]{
\label{fig:examples-b}
\includegraphics[width=0.5\columnwidth]{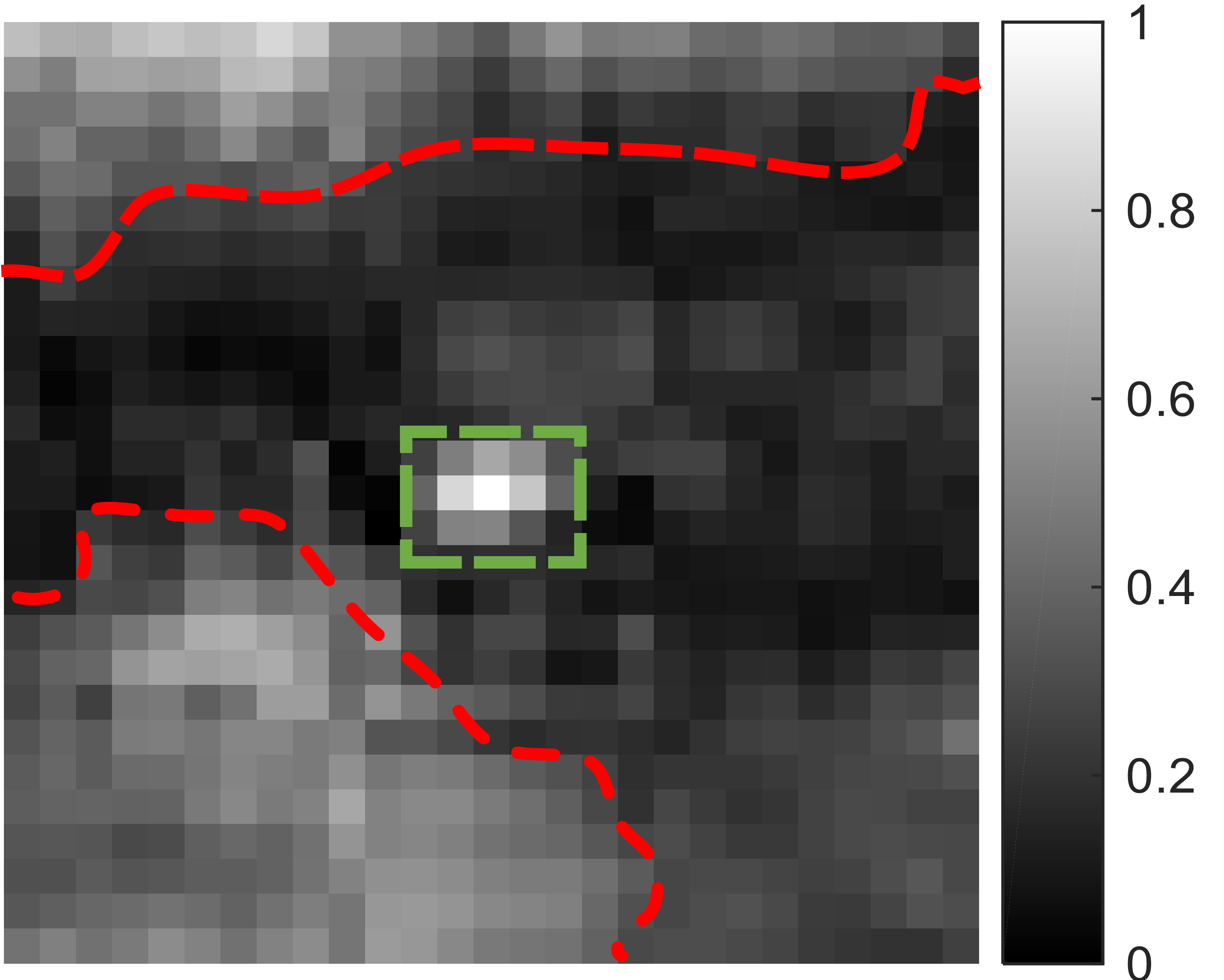}}\hfill
\subfloat[]{
\label{fig:examples-c}
\includegraphics[width=0.5\columnwidth]{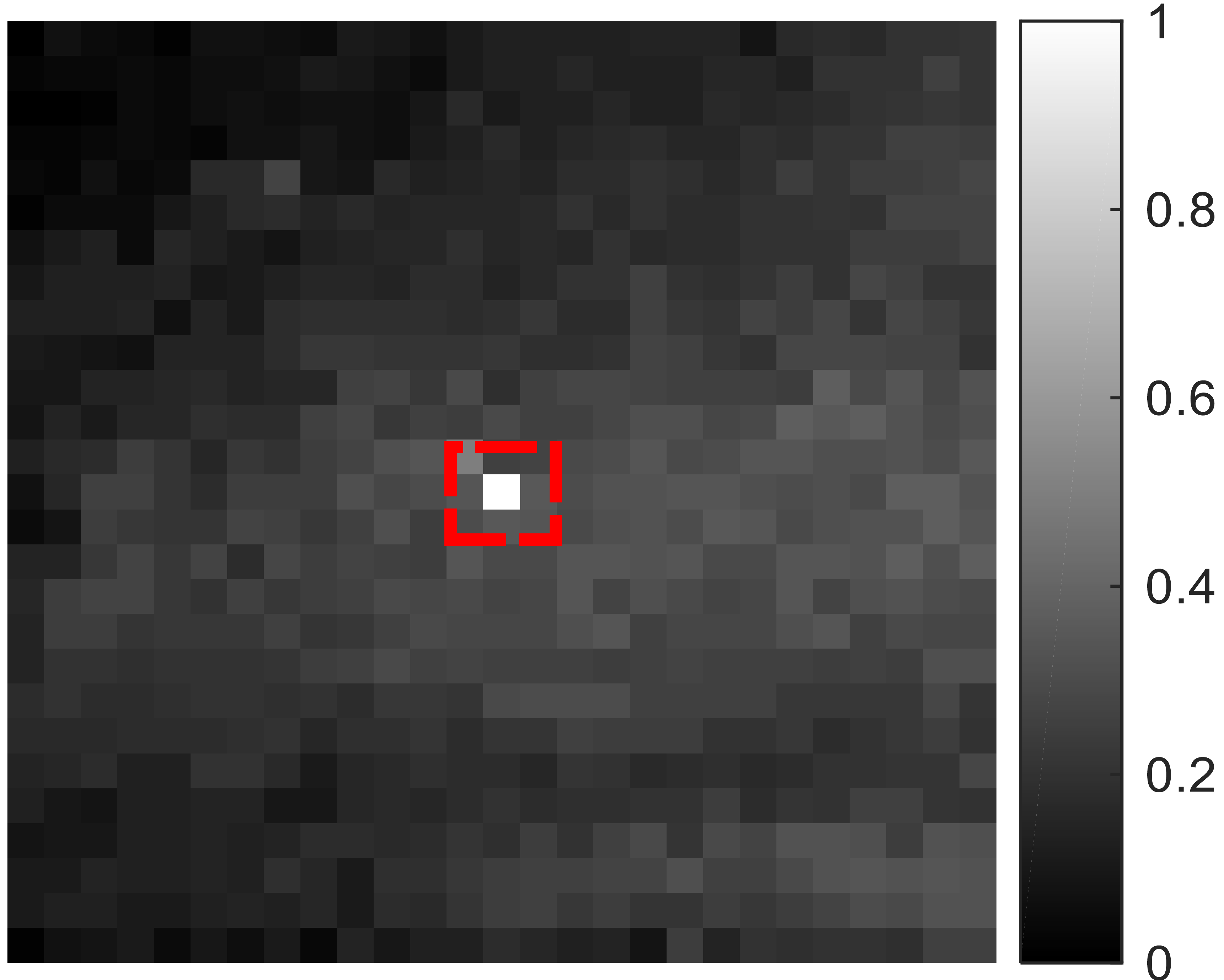}}
\subfloat[]{
\label{fig:examples-d}
\includegraphics[width=0.5\columnwidth]{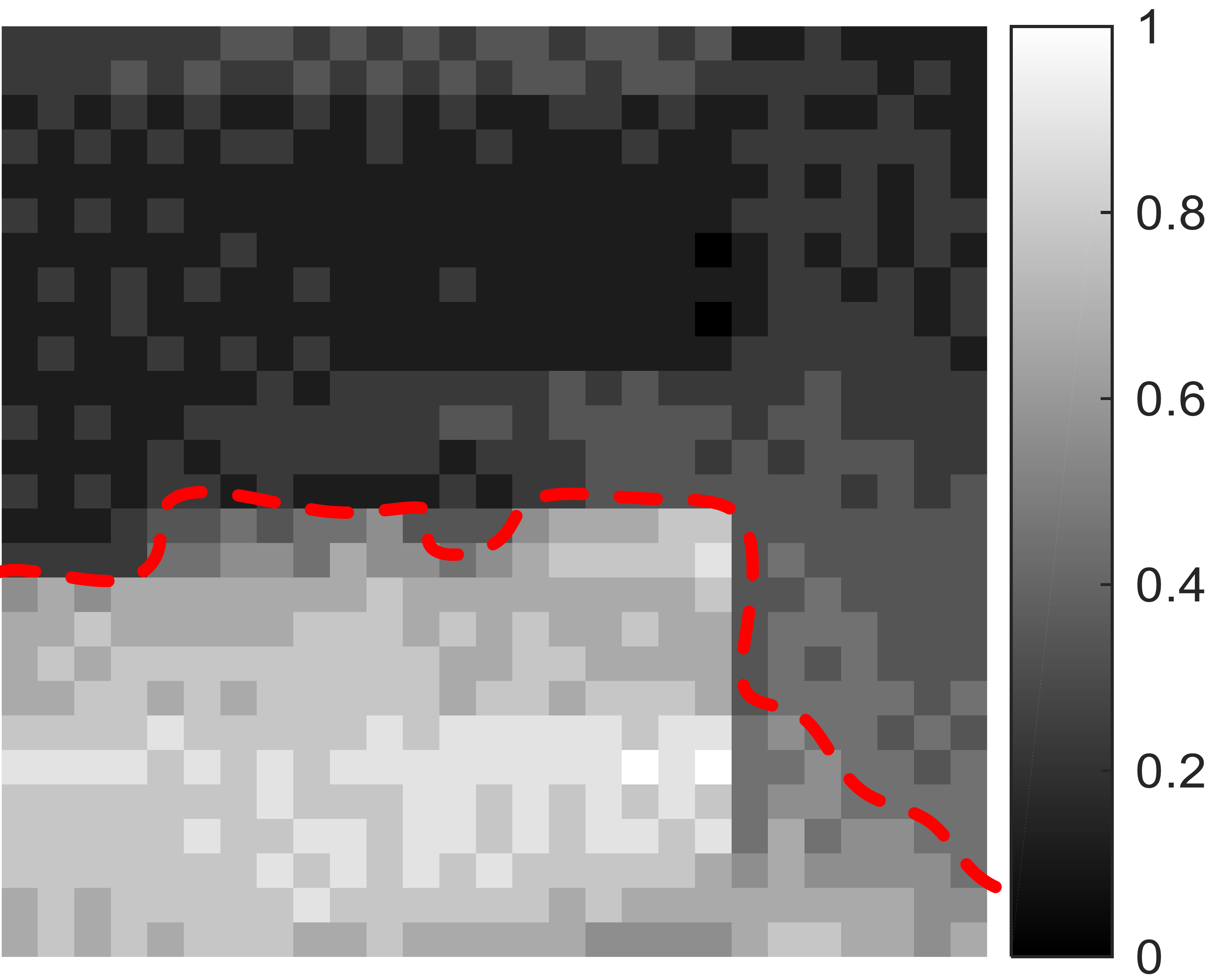}}
\caption{Practical image patches of (a-b) small targets, (c) PNHB and (d) strong edge of clutters. The green dashed rectangle indicates the small targets, the red dashed rectangle indicates the PNHB and the red dashed curve indicates the strong edges. Best viewed in color.}
\label{fig:examples}
\end{figure}

Recall \eqref{eq:HMERW stationary} that the stationary distribution of the HMERW is dependent on the weight matrix \textbf{W}. Thus, a well-defined weight matrix \textbf{W} is crucial for our small target detection model. Note that the edge weight of the MERW is quantified by node dissimilarity, which indicates target nodes that show considerable dissimilarity owns high probabilities to be visited. The intention of designing a proper \textbf{W} is to measure the contrast consistency and regional compactness characteristics of a small target. There are two common choices for defining the weight of nodes $v_i$ and $v_j$, that is, the Euclidean distance:
\begin{eqnarray}\label{eq:Euclidean-W}
\textbf{W}^{\text{Euclidean}}_{ij}=\|\textbf{I}(v_i)-\textbf{I}(v_j)\|^2,
\end{eqnarray}
and the Gaussian distance:
\begin{eqnarray}\label{eq:Gaussian-W}
\textbf{W}^{\text{Gaussian}}_{ij}=\textbf{e}^{\alpha\|\textbf{I}(v_i)-\textbf{I}(v_j)\|^2},\alpha\in\mathbb{R}.
\end{eqnarray}

An underlying assumption of assigning the Euclidean distance or Gaussian distance to the weight matrix is that the pixels of a target have relatively higher or lower intensities than their adjacent pixels. However, this assumption is not true if there exists strong interferences in the infrared images, for example, strong edges of clutters and PNHBs. Fig. \ref{fig:examples} shows some practical image patches of small targets and strong interferences that reveal similar local characteristics. To distinguish them, we design a specific weight matrix for the HMERW.

\begin{figure}[htbp]
\centering
\includegraphics[]{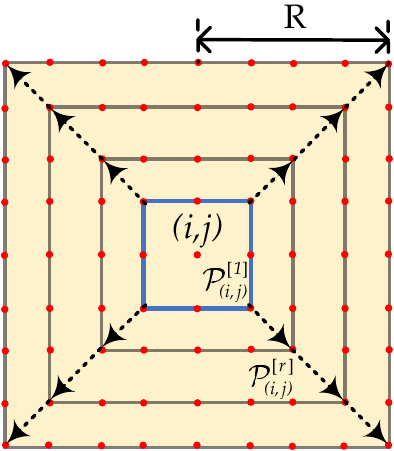}
\caption{Diffusion structure of image patch $\mathcal{P}_{(i,j)}$.}
\label{fig:diffusion-structure}
\end{figure}

As seen from Fig. \ref{fig:examples}, although the marked pixels, including small targets, PNHB and strong edges, exhibit remarkable local contrast characteristic, there are still some major differences among them. To be specific, small targets in Fig. \ref{fig:examples-a} and Fig. \ref{fig:examples-b} spread throughout the local regions of more than $3\times 3$ pixels, whereas the PNHB in Fig. \ref{fig:examples-c} occupies fewer pixels. This is called regional compactness characteristic. As shown  in Fig \ref{fig:examples-b} and Fig. \ref{fig:examples-d}, intensities of the strong edges marked with red dashed curves drop (or rise) suddenly along specific directions. By contrast, intensities of small targets gradually vary inside out along all directions. We call it the contrast consistency characteristic. Inspired by the regional compactness and contrast consistency characteristics of a small target, a specific weight matrix is designed to enhance small targets and suppress interferences. 

Given a pixel $(i,j)$ in the filtered image $\textbf{I}\in\mathbb{R}^{m\times n}$, build an image patch with size of $(2R+1)\times (2R+1)$ centered at $(i,j)$ as $\mathcal{P}_{(i,j)}=\{\mathcal{P}^{[r]}_{(i,j)}|r=0,1,\dots,R\}$, where 
$$
\mathcal{P}^{[r]}_{(i,j)}=\{(u,v)\ |\ max(|u-i|,|v-j|)=r\}.
$$
A schematic diagram of image patch construction is shown in Fig. \ref{fig:diffusion-structure}. Specially, $\mathcal{P}^{[0]}_{(i,j)}$ consists of only one pixel $(i,j)$.

Recall that $\mathcal{V}=\{v_1,\dots,v_N\}$ is the set of nodes corresponding to the pixels of \textbf{I}, we define a function:
\begin{eqnarray}\label{eq:similar}
\text{Simi}(v_i,r)=\{v_s\ |\ |\text{Imean}(\mathcal{P}^{[1]}_{v_i})-\text{I}(v_s)|=\text{min},v_s\in\mathcal{P}^{[r]}_{v_i}\},
\end{eqnarray}
where  Imean($\cdot$) is a function that calculates the mean intensity of the input nodes, and \emph{min} denotes the minimal value of $|\text{Imean}(\mathcal{P}^{[1]}_{v_i})-\textbf{I}(v_s)|$ for any $v_s\in\mathcal{P}^{[r]}_{v_i}$. In essence, function $\text{Simi}(v_i,r)$ returns a node $v_s$ of the $r$-th sub-patch $\mathcal{P}^{[r]}_{v_i}$, with intensity $\textbf{I}(v_s)$ most ``similar'' to $\text{Imean}(\mathcal{P}^{[1]}_{v_i})$.
For any node $v_{i}\in\mathcal{V}$, the weight between node $v_i$ and node $v_j\in\mathcal{V}\backslash\{v_i\}$ is defined as follow,
\begin{eqnarray}\label{eq:our-weight}
\textbf{W}^{\text{RCCC}}_{ij}=\left\{
\begin{aligned}
&|\text{Imean}(\mathcal{P}^{[1]}_{v_i})-\textbf{I}(v_j)|*\frac{\text{Imean}(\mathcal{P}^{[r]}_{v_i}\backslash\{v_j\})}{\text{Imean}(\mathcal{P}^{[r]}_{v_i})},  \\
& \text{if }v_j\in\{\text{Simi}(v_i,r)\ | \ r=2,\dots,R\}\\
&0,                                         \text{otherwise}
\end{aligned}\right.
\end{eqnarray}
where $\mathcal{P}^{[r]}_{v_i}\backslash\{v_j\}$ means excluding node $v_j$ from the sub-patch $\mathcal{P}^{[r]}_{v_i}$. The superscript ``RCCC'' of \textbf{W} implies that the designed weight matrix manages to describe regional compactness and contrast consistency of a small target.

The way of weight definition of \eqref{eq:our-weight} has following advantages. First, only neighboring connection of nodes is considered in accordance with the local characteristics of a small target. Second, instead of using $\textbf{I}(v_i)$ directly, we use the mean intensity of the sub-patch $\mathcal{P}^{[1]}_{v_i}$ to compute intensity difference between nodes $v_i$ and $v_j$. This is helpful for emphasizing regional compactness of a small target and suppressing PNHBs. Moreover, in the first term of \eqref{eq:our-weight}, at each sub-patch, only the node with intensity  most close to $\textbf{I}(v_i)$  is used to compute intensity difference. This helps to suppress strong edges since the intensity variation of strong edges is directional while that of a small target is omni-directional. As for the second item of \eqref{eq:our-weight}, it evaluates the intensity contribution of node $v_j$ to its sub-patch, the more it contributes, the smaller the weight is. In this way, the  intends to emphasize the intensity smoothness of sub-patch $\mathcal{P}_{v_i}^{[r]}$, which in turn enhances the contrast consistency of a small target.

\begin{figure}[htbp]
\centering 
\subfloat[]{
\label{fig:examples-b-norm2}
\includegraphics[width=0.5\columnwidth]{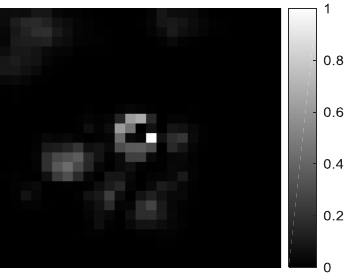}}
\subfloat[]{
\label{fig:examples-c-norm2}
\includegraphics[width=0.5\columnwidth]{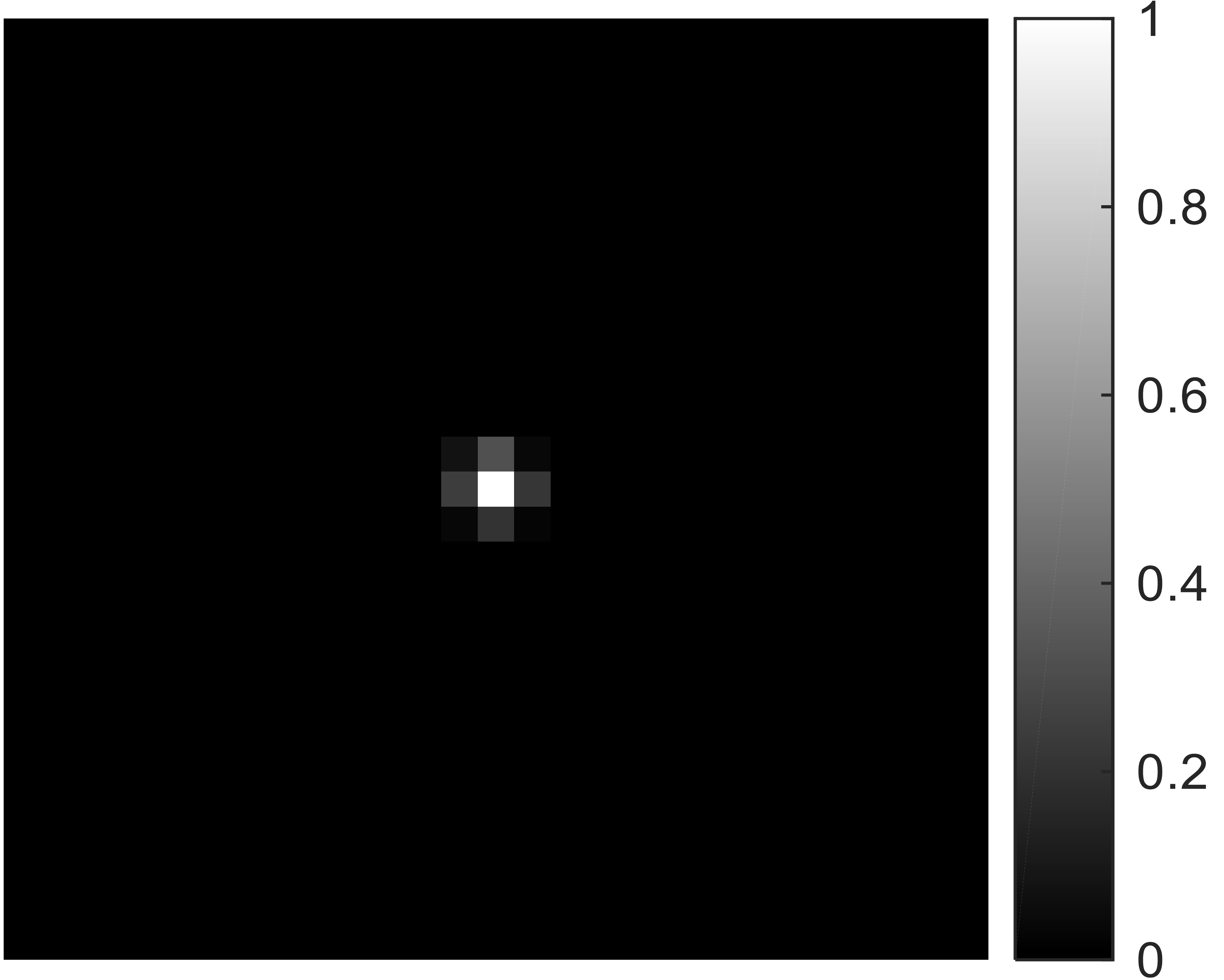}}\hfill
\subfloat[]{
\label{fig:examples-b-MERW}
\includegraphics[width=0.5\columnwidth]{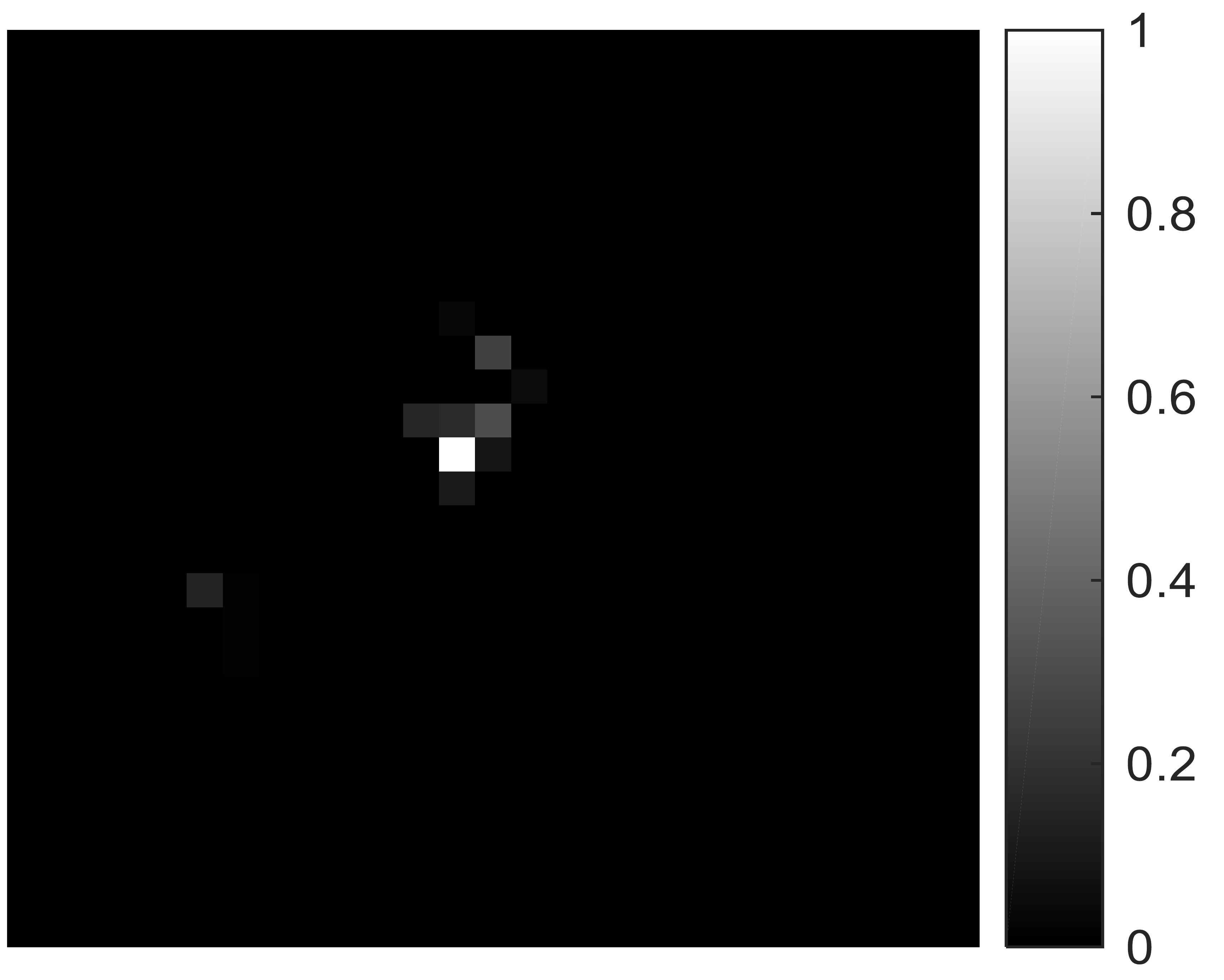}}
\subfloat[]{
\label{fig:examples-c-MERW}
\includegraphics[width=0.5\columnwidth]{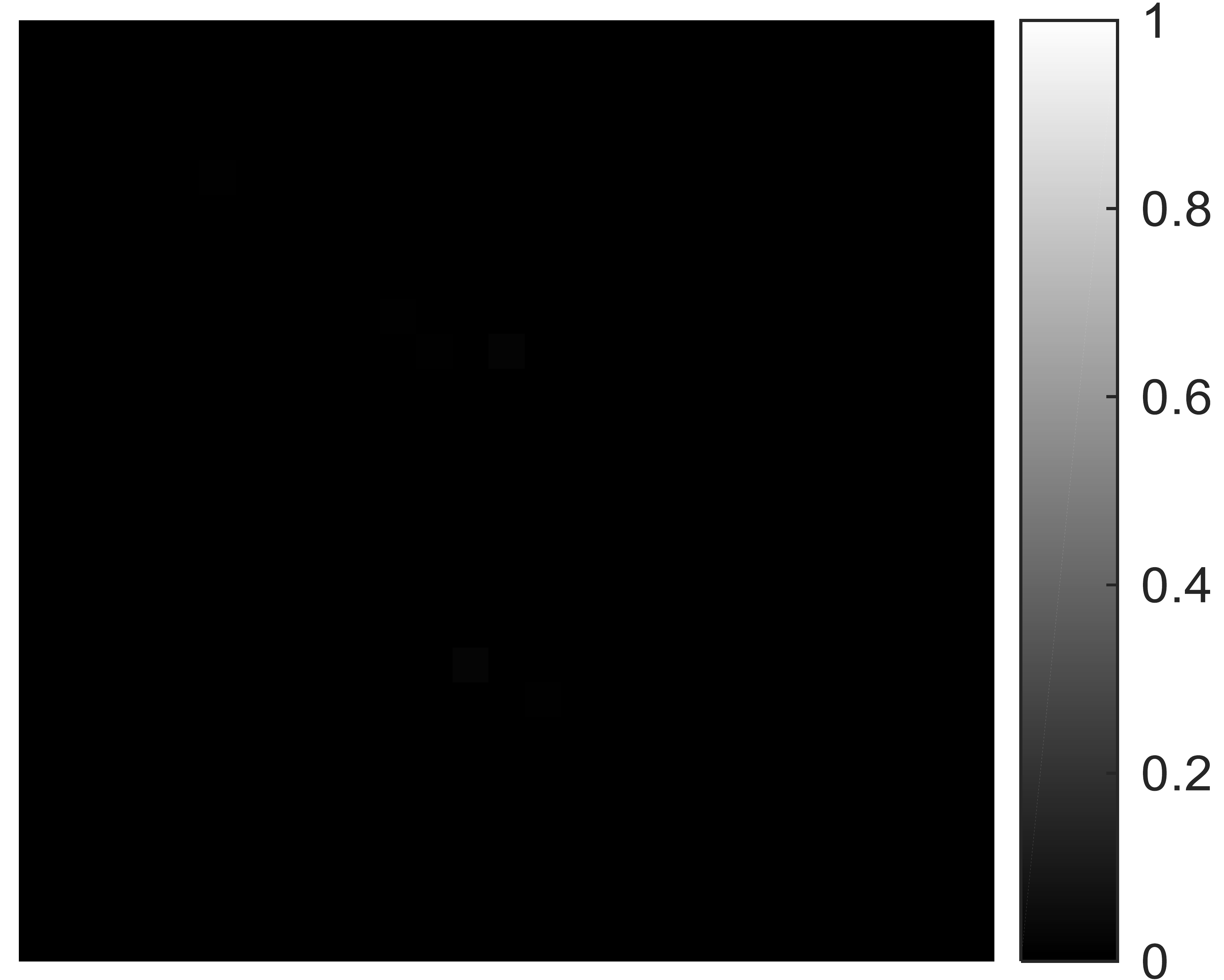}}
\caption{Stationary distribution maps of the HMERW tested on (a) Fig. \ref{fig:examples-b} with $\textbf{W}^\text{Euclidean}$, (b) Fig. \ref{fig:examples-c} with $\textbf{W}^\text{Euclidean}$, (c) Fig. \ref{fig:examples-b} with $\textbf{W}^\text{HMERW}$ and (d) Fig. \ref{fig:examples-c} with $\textbf{W}^\text{HMERW}$.  (Note that the grayscale of (d) is close to zero.)}
\label{fig:examples-pi}
\end{figure}

Note that the weight matrix $\textbf{W}^\text{RCCC}$ is asymmetric in most cases. To be specific, it is probable that $\text{Simi}(v_i,r)=v_j$ while $\text{Simi}(v_j,r)\ne v_i$, which results in $\textbf{W}^\text{RCCC}_{ij}\ne\textbf{W}^\text{RCCC}_{ji}$. However, the HMERW requires a symmetric weight matrix. To solve this problem, we redefine a symmetric version of $\textbf{W}^\text{RCCC}$ for the HMERW as follows,
\begin{eqnarray}\label{eq:symmetric-our-weight}
\textbf{W}^\text{HMERW}=\frac{\textbf{W}^\text{RCCC}+(\textbf{W}^\text{RCCC})^\text{T}}{2}.
\end{eqnarray}
Then, the stationary distribution $\boldsymbol{\pi}^\text{HMERW}$ of the HMERW can be obtained according to \eqref{eq:HMERW stationary}. To illustrate the effectiveness of the designed weight matrix for small target enhancement and interference suppression, we respectively test the HMERW with $\textbf{W}^\text{Euclidean}$ (as seen in \eqref{eq:Euclidean-W}) and $\textbf{W}^\text{HMERW}$ on practical image patches of Fig. \ref{fig:examples-b} and Fig. \ref{fig:examples-c}, and present their stationary distribution maps in Fig. \ref{fig:examples-pi}. It can be seen from Fig. \ref{fig:examples-pi} that the designed weight matrix can help the HMERW robustly suppress strong edges (Fig. \ref{fig:examples-b-MERW}) and PNHB (Fig. \ref{fig:examples-c-MERW}), while the $\textbf{W}^\text{Euclidean}$-version HMERW mistakenly enhances them. In addition, both versions of the HMERW perform well in small target enhancement. In the following sections, we refer to the stationary distribution map of the $\textbf{W}^\text{HMERW}$-version HMERW as HMERW map.

\begin{figure}[htbp]
\centering
\includegraphics[width=\columnwidth]{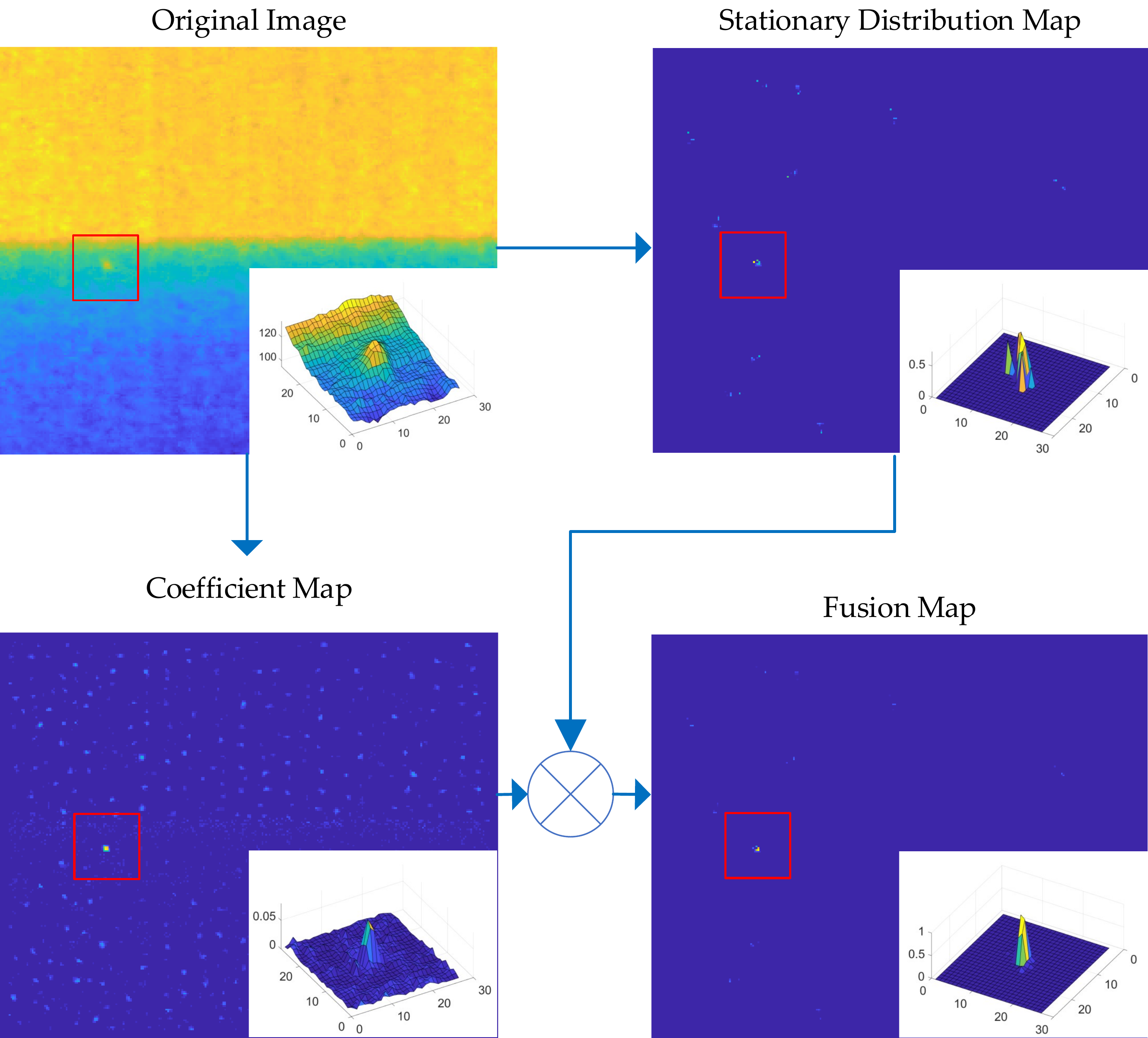}
\caption{Schematic diagram of map fusion process. The bottom right of each image is a 3D mesh of the zoom-in patch of a small target. (Pseudo-color images are transformed from gray-scale images.)}
\label{fig:image-fusion}
\end{figure}

\subsection{Fusion technique}
\label{sec:coefficient}
Although most clutters have been eliminated in the HMERW map, some trivial pixels around the real target are also mistakenly enhanced, as shown in the zoom-in patch of the HMERW map in Fig. \ref{fig:image-fusion}. This happens due to our definition of the symmetric weight matrix $\textbf{W}^\text{HMERW}$, which assumes that a random walker has equivalent chance of walking from the target node to its neighboring nodes and walking from neighboring nodes to the target node. To fix it, we take advantage of the asymmetric weight matrix $\textbf{W}^\text{RCCC}$ and construct a coefficient map, and then merge it with the HMERW map. Fig. \ref{fig:image-fusion} shows the schematic diagram of the fusion process.

As mentioned before, the design principle of  $\textbf{W}^\text{RCCC}$ is to emphasize regional compactness and contrast consistency characteristics of small targets, which suggests that the degrees ($d_i=\sum_j\textbf{W}^\text{RCCC}_{ij}$) of nodes corresponding to small targets have larger values than the degrees of their neighboring nodes. From this point of view, we design a coefficient vector $\textbf{c}\in\mathbb{R}^{1\times N}$ as
\begin{eqnarray}\label{eq:weighted-map}
\textbf{c}_i=\frac{1}{R-1}\sum_{v_j\in\mathcal{V}}\textbf{W}^\text{RCCC}_{ij},
\end{eqnarray}
where $R$ is the scale parameter of image patch $\mathcal{P}_{(i,j)}$, as shown in Fig. \ref{fig:diffusion-structure}. An example of a coefficient map transformed from a practical infrared image is shown in the bottom left of Fig. \ref{fig:image-fusion}, from which we can see that the trivial pixels of the target neighborhood are flattened, however, the strong interferences are not well suppressed. Combining the stationary distribution $\boldsymbol{\pi}^\text{HMERW}$ and the coefficient vector $\textbf{c}$, we have
\begin{eqnarray}\label{eq:fusion-map}
\textbf{f}=\boldsymbol{\pi}^\text{HMERW}\otimes\textbf{c},
\end{eqnarray}
where $\otimes$ indicates element-wise multiplication. A fusion map \textbf{F} obtained by reshaping \textbf{f} into a matrix with same size of \textbf{I} is shown in the bottom right of Fig. \ref{fig:image-fusion}, in which the small target is so salient that can be readily segmented from the background.

\subsection{Small target detection algorithm based the HMERW}
In the fusion map \textbf{F}, small targets are easy to be extracted out from background. Here, an adaptive threshold is used,
\begin{eqnarray}\label{eq:threshold}
T=\text{mean}(\textbf{F})+\lambda*\text{std}(\textbf{F}),
\end{eqnarray}
where $\lambda$ is a constant, $\text{mean}(\cdot)$ and $\text{std}(\cdot)$ compute mean and standard deviation of the input matrix, respectively. Extensive experiments suggest that selecting $\lambda$ from an interval [5,20] provides good detection performance.

Based on the aforementioned preparations, a detailed algorithm of the proposed small target detection method is given in Algorithm \ref{alg:framework}. In Algorithm \ref{alg:framework}, the size of the input image is $N=m\times n$, the optimal choice of parameters $R$ and $K$ will be discussed in Section \ref{sec:experiment}.
\begin{algorithm}
\caption{Proposed small target detection method based on the HMERW}
\label{alg:framework}
\begin{algorithmic}[1]
\Require 
Input infrared image $\textbf{I}_0$, image patch parameter $R$, eigenvalue number $K$, threshold parameter $\lambda=10$.
\Ensure 
Target position $(x,y)$.
\State Initialization: $\textbf{W}=\textbf{0}_{N\times N}$, $\boldsymbol{\pi}=\textbf{0}_{1\times N}$, $\textbf{c}=\textbf{0}_{1\times N}$, $\textbf{f}=\textbf{0}_{1\times N}.$
\State Perform mean filtering on the input image with a $2\times2$ template and obtain a filtered image $\textbf{I}$;
\State Build a graph structure for $\textbf{I}$ as $\mathcal{G}=(\mathcal{V},\mathcal{E})$;
\For{$i=1:N$}
	\For{$r=1:R$}
			\State Construct a sub-patch $\mathcal{P}^{[r]}_{v_i}$ for node $v_i$;
		\If{$r==1$}
			\State $\mu^{[1]}_{v_i}=\text{Imean}(\mathcal{P}^{[1]}_{v_i})$;
		\Else
			\State Search $v_j=\text{Simi}(v_i,r)$ according to \eqref{eq:similar};
			\State $\tilde{\mu}^{[r]}_{v_i}=\text{Imean}(\mathcal{P}^{[r]}_{v_i}\backslash\{v_j\})$;
			\State $\mu^{[r]}_{v_i}=\text{Imean}(\mathcal{P}^{[r]}_{v_i})$;
			\State $\textbf{W}_{ij}=|\mu^{[1]}_{v_i}-\textbf{I}(v_j)|*\tilde{\mu}^{[r]}_{v_i}/\mu^{[r]}_{v_i}$;
		\EndIf
		
	\EndFor
\EndFor
\State Compute a coefficient vector $\textbf{c}$ according to \eqref{eq:weighted-map};
\State Obtain a symmetric weight matrix: $\textbf{W}=(\textbf{W}+\textbf{W}^\text{T})/2$;
\State Perform eigenvalue decomposition on $\textbf{W}$ and obtain top $K$ eigenvalues $\{\lambda_1,\dots,\lambda_{K}\}$ and top $K$ eigenvectors $\{\boldsymbol{\psi}^{(1)},\dots,\boldsymbol{\psi}^{(K)}\}$;
\State Compute the stationary distribution $\boldsymbol{\pi}$ according to \eqref{eq:HMERW stationary};
\State Compute a fusion vector $\textbf{f}$ according to \eqref{eq:fusion-map};
\State Reshape $\textbf{f}$ and obtain a fusion map $\textbf{F}$ for $\textbf{I}$;
\State Compute threshold $T$ according to \eqref{eq:threshold};
\State Output pixels $(x,y)$ with $\textbf{F}(x,y)>T$.

\end{algorithmic}
\end{algorithm}

\begin{table*}[htbp]
\centering
\caption{\\INFORMATION OF THE TEST DATA SETS.}
\begin{tabular}{l l l l l l l l l}
\toprule
                   &\#Frames &\#Targets &Frame Size &Target Category &Target Size  &Background Type &Image quality\\
\midrule
S1   &30 &44 &256$\times$200  &Airplane &6$\times$4  &Heavy cloud &Heavy noise\\
\midrule
S2   &300  &300 &320$\times$250 &Airplane &5$\times$3 to 5$\times$7  &Sky &Heavy pepper noise and many dead pixels\\
\midrule
S3   &100  &100 &320$\times$240 &Bird    &4$\times$4 to 9$\times$7 &Sky-ground   &Overexposure and dark margins\\
\midrule
SC  &29  &34 &125$\times$125  &Cars, ships, etc.  &3$\times$3 to 9$\times$7  &Sea, road, etc.  &Complex exposure conditions\\             
\bottomrule
\end{tabular}
\label{tb:dataset}
\end{table*}

\section{Contributions and Novelties}\label{sec:novelty}
There are several contributions and novelties of this paper can be described as follows. First, we explore the feasibility of the primal MERW for measuring the global uniqueness of a small target.  
Global uniqueness is a crucial characteristic of a small target, however, only few studies exploit this characteristic for small targets. Although random walk model has been successfully applied to small target detection in \cite{rs10122004}, it lacks explicit expression of involving global awareness during the Markov process. By comparison, we find that the MERW model is more expressive in acquiring global information during the random process, which can be perceived from its transition probability of \eqref{eq:MERWprob}. Simulation experiments on synthetic dataset validates the feasibility of applying MERW to enhance the global uniqueness of a small target. In addition, practical single small target detection experiments based on MERW+$\textbf{W}^\text{Euclidean}$, which are presented in Section \ref{sec:ablation}, further demonstrate the effectiveness of MERW for enhancing global uniqueness.

Second, we analyze the limitation of the primal MERW and attest that the limitation is caused by insufficient graph projection. 
Although MERW model has been transferred from field of thermal dynamics to many other scenarios, e.g., saliency detection and link prediction \cite{10.1145/2063576.2063741}, none of these studies intends to analyze the limitation of MERW  and modify it. We find that the MERW model is strongly biased to nodes with dominant information, and further investigate that this limitation results from the insufficient weight matrix reconstruction, which corresponds to insufficient graph projection, as described in Section \ref{sec:HMERW}. Derived from this observation, we develop a theory of graph decomposition. The graph decomposition theory ensures minimal information loss after projecting a graph along a specific direction onto a corresponding subspace. Several benefits can be gained from the graph decomposition theory. First, we are able to analyze the decomposed graph from different perspectives through graph projection, as sub-graphs spanned on different subspaces contain specific features of the primal graph. Second, analysis of sub-graphs is independent of  each other. Therefore, we can readily apply this theory to various graph-based models without any modification, e.g., graph neural network.

Third, a hierarchical version of MERW (HMERW) model is proposed based on the graph decomposition theory, and a specific weight matrix is designed to enhance contrast consistency and regional compactness characteristics of small targets. 
To authors' best knowledge, this is the first study to modify the primal MERW model to satisfy practical applications. The HMERW overcomes the limitation of strong bias of the primal MERW and succeeds in separating multiple anomalies from homogeneous nodes. This advantage of multiple anomalies enhancement is derived from the mechanism of graph decomposition and our specifically designed weight matrix. Likewise, we can transform the proposed HMERW to existing MERW-based models for various tasks without much effort.

Finally and most importantly, a small target detection method is designed based on the HMERW.
Multiple small targets detection is an important task in IRST systems. Motivated by the advantages of the HMERW in representing global uniqueness and detecting multiple anomalies, we proposed a small target detection method based on the HMERW. Our intention is to incorporate both local and global characteristics of a small target into the HMERW model. In the proposed method, a coefficient map is presented based on the proposed weight matrix and used to fuse the output map of the HEMRW model. Extensive experiments on real data sets demonstrate that the proposed HMERW-based detection method outperforms state-of-the-art small target detection methods in target enhancement, background suppression and detection performance. In particular, the state-of-the-art methods fail in detecting multiple small targets, while our HMERW-based method can achieve satisfactory detection performance.

\section{Experimental results}\label{sec:experiment}
In this section, we first introduce the experimental data sets and evaluation metrics for small target detection. Next, detailed analysis of the parameters involved in the proposed small target detection method is provided. After that, how the proposed HMERW-based method achieves multiple small targets detection is analyzed. Then, each component included in the proposed is discussed. Moreover, ablation studies are implemented to demonstrate the effectiveness of each component. Finally, comprehensive comparison between the proposed method and baseline methods are given. All experiments are performed on a computer with a 3.6 GHz Intel core i7 CPU and 8GB RAM, and the code is implemented using MATLAB R2018a.

\subsection{Experimental data sets and evaluation metrics}
More than 450 practical infrared images including three consecutive sequences and a collection of single-frame infrared images are used for experiments. Detailed information of the data sets is listed in Table \ref{tb:dataset}. For simplicity, we refer to the consecutive sequences and the image collection as abbreviations S1, S2, S3 and SC, respectively. In particular, SC contains single-frame infrared images photographed under various scenes, e.g., sailing boat on the sunlit lake, running cars on the country road and flying airplane under dusk. Therefore, SC is a good data set for validating robustness of  small target detection methods.

Generally, small target detection methods can be evaluated in three aspects, i.e., target enhancement, background suppression and detection accuracy. 
Following \cite{rs10122004}, we use local contrast gain (\emph{LCG}) for target enhancement evaluation, and background suppression factor (\emph{BSF}) for background suppression evaluation. \emph{LCG} is defined as follows,
\begin{eqnarray}\label{eq:LCG}
\emph{LCG}=\frac{(\mu_\mathcal{T}(\textbf{O})-\mu_\mathcal{B}(\textbf{O}))/(\mu_\mathcal{T}(\textbf{O})+\mu_\mathcal{B}(\textbf{O})+\epsilon)}{(\mu_\mathcal{T}(\textbf{I})-\mu_\mathcal{B}(\textbf{I}))/(\mu_\mathcal{T}(\textbf{I})+\mu_\mathcal{B}(\textbf{I}))+\epsilon},
\end{eqnarray}
where $\textbf{I}$ is the original image,  $\textbf{O}$ is the output map of a small target detection algorithm, $\mu_\mathcal{T}(\cdot)$ and $\mu_\mathcal{B}(\cdot)$ denote the mean grayscale of the target region and background region of the input map, respectively, $\epsilon$ is a small constant to avoid meaningless computation. Here, we set $\epsilon$ to $10^{-7}$. \emph{BSF} is defined by
\begin{eqnarray}\label{eq:BSF}
\emph{BSF}=\frac{\sigma(\textbf{O})}{\sigma(\textbf{I})+\epsilon},
\end{eqnarray}
where $\sigma(\cdot)$ calculates the standard deviation of the input map.

According to \cite{myownpaper}, the precision and recall (PR) curve is more suitable than the receiver operating characteristic (ROC) curve for evaluating detection performance of small target detection methods since there are much more negative objects (pixels of background) than positive objects (pixels of small targets). PR curve is a graphic plot of precision ($P$) versus recall ($R$), of which $P$ and $R$ are defined by
\begin{eqnarray}\label{eq:PR}
\begin{aligned}
P=\frac{N_{\text{tp}}}{N_{\text{tp}}+N_{\text{fp}}},\\
R=\frac{N_{\text{tp}}}{N_{\text{p}}},
\end{aligned}
\end{eqnarray}
where $N_\text{tp},N_\text{fp}$ and $N_\text{p}$ denote the number of true detections, false detections and ground truth of the data sets, respectively.
For clarity, the output maps obtained by different small target detection methods are normalized to 8-bit grayscale images. For a particular data set, precision and recall are computed by varying the segmentation threshold $T$ from 0 to 255 with an interval of 1. Then, the PR curves can be generated. Besides, a detection result is considered as a positive detection if the Chebyshev distance (maximum coordinate distance along any coordinate dimension) between the detection result and the center of actual target is less than 4 pixels. Another used quantitative metric for evaluating detection performance is the area under precision-recall curve (\emph{AUPR}), which ranges from 0 to 1 \cite{Davis:2006:RPR:1143844.1143874}.

\subsection{Analysis of crucial parameters}\label{eq:analysis}
There are two crucial parameters in our method, that is, the image patch scale $R$ in Fig. \ref{fig:diffusion-structure} and the number of eigenvalues $K$ in \eqref{eq:HMERW stationary}. 
For quantitative analysis, one parameter is analyzed by holding the other one fixed. By default, $R$ and $K$ are respectively set to 5 and 30 in the following experiments.

\subsubsection{Analysis of $R$}\label{sec:analysis-R}
As shown in Fig. \ref{fig:diffusion-structure}, the image patch scale $R$ directly affects the construction of the weight matrix for the HMERW, and then further affects the output map and detection result. According to \eqref{eq:our-weight}, the parameter $R$ determines the neighborhood size of pixels to describe their local characteristics. To properly emphasize the local characteristics of a small target, the best choice for $R$ is a slightly larger value than the actual size of a small target. To verify it, we vary $R$ from 2 to 7 with an interval of 1 and test it on the data sets. Table \ref{tb:AUPR-R} reports the evaluation results of \emph{LCG}, \emph{BSF} and \emph{AUPR} for different values of $R$, from which we can see that the proposed method achieves good \emph{LCG} and \emph{BSF} results in spite of the change of $R$. This indicates the robustness of the proposed HMERW to small target enhancement and background suppression. 
As for the results of \emph{AUPR}, the optimal choice of $R$ is different for different data sets. If the parameter $R$ is too small, some small-scale pseudo targets will be mistakenly enhanced, for example, the consecutive dead pixels (consecutive damaged pixels with intensities of zeros) in the data set S2. However, it will involve unnecessary computation if $R$ is too large. In summary, for each data set, setting the value of $2R+1$ to around 3 larger than the actual target size (as reported in Table \ref{tb:dataset}) yields the best \emph{AUPR} result. As defined by Society of Photo-Optical Instrumentation Engineers (SPIE), a small target occupies less than 80 pixels in an infrared image \cite{8723140}. Therefore, we set $R=5$ by default in practical application.

\begin{table}[h]
\caption{\emph{LCG}, \emph{BSF} and \emph{AUPR} results of testing the proposed method on the data sets with different values of $R$.
}
\label{tb:AUPR-R}
\begin{threeparttable}
\centering
\begin{tabular}{l l l l l l l l}
\toprule
	&		&$R=2$ &$R=3$ &$R=4$  &$R=5$ &$R=6$ &$R=7$\\
\midrule
S1  &$\overline{\emph{LCG}}$    &7.84  &7.62  &7.87  &$\textbf{8.89}^{*}$  &7.53  &8.11\\
	&$\overline{\emph{BSF}}$    &14.37 &\textbf{19.87}  &19.16  &18.06  &17.77  &17.53\\
	&\emph{AUPR} &0.89  &0.94  &0.96   &\textbf{0.99}  &0.96  &0.97\\
\midrule
S2  &$\overline{\emph{LCG}}$    &7.75  &8.73  &\textbf{8.86}   &8.75  &8.69  &8.66\\
	&$\overline{\emph{BSF}}$	 &\textbf{45.22}  &44.19  &43.04 &42.53  &42.17  &42.07\\
	&\emph{AUPR} &0.26   &0.71  &0.87  &\textbf{0.93}  &0.93 &0.93\\
\midrule
S3  &$\overline{\emph{LCG}}$   &9.55   &9.66  &9.65  &\textbf{9.78}  &9.36   &9.38\\
	&$\overline{\emph{BSF}}$    &50.86  &60.17  &\textbf{61.15}  &57.67  &54.32  &52.72\\
	&\emph{AUPR} &0.85   &0.99  &\textbf{1.00}  &\textbf{1.00}   &\textbf{1.00}  &\textbf{1.00}\\
\midrule
SC &$\overline{\emph{LCG}}$    &8.99  &\textbf{8.99}  &8.95  &8.93  &8.94  &8.46\\
	&$\overline{\emph{BSF}}$    &17.68 &\textbf{17.91}  &16.69  &16.12  &15.56  &15.40\\
	&\emph{AUPR} &0.99  &\textbf{1.00}  &\textbf{1.00} &\textbf{1.00}   &0.99   &0.96\\
\bottomrule
\end{tabular}
\begin{tablenotes}
\footnotesize
        \item[*] The bolder data means the maximum of indicators.
\end{tablenotes}
\end{threeparttable}
\end{table}

\begin{figure*}[htbp]
\centering 
\subfloat[]{
\label{fig:K-LCG}
\includegraphics[width=0.33\textwidth]{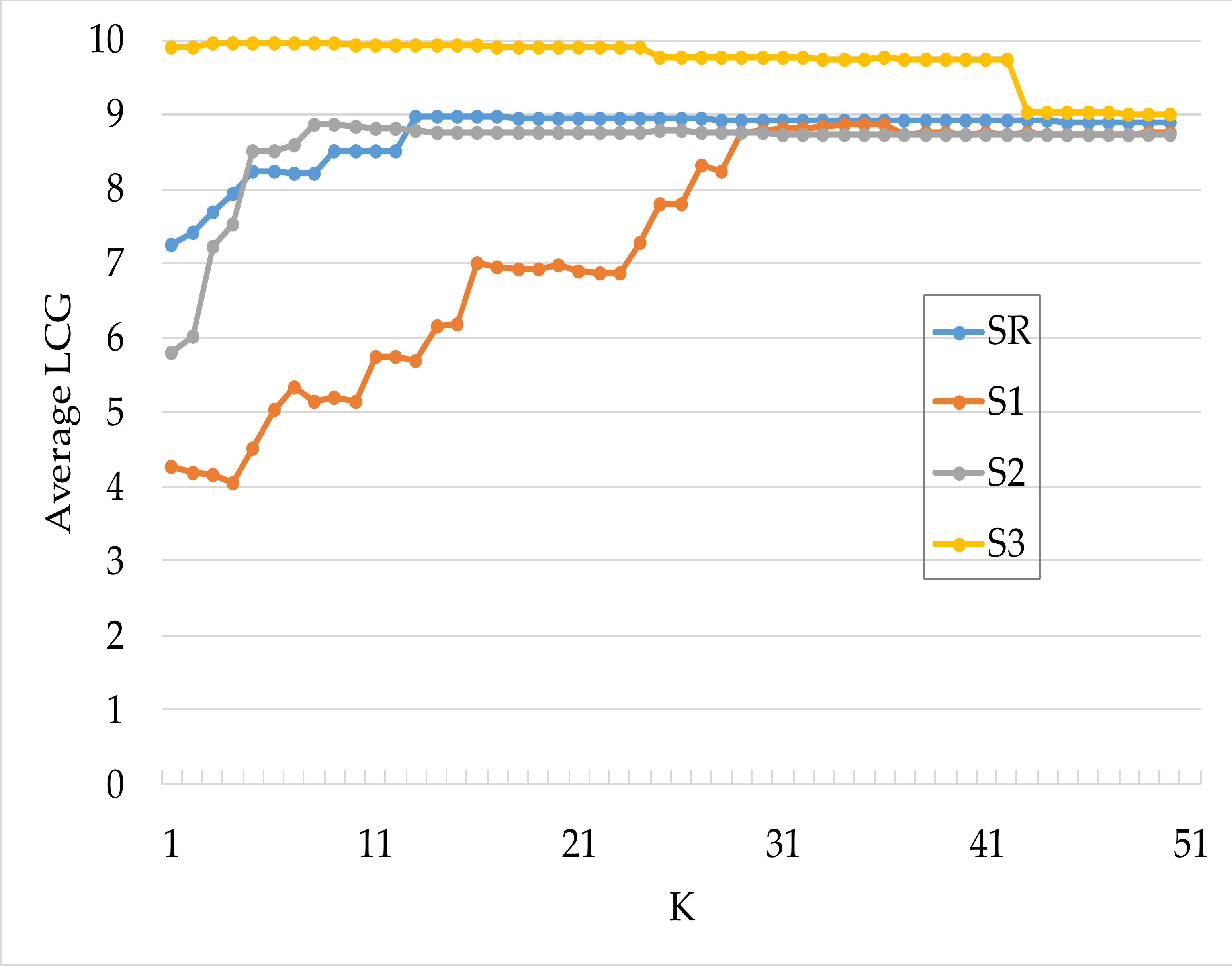}}
\subfloat[]{
\label{fig:K-BSF}
\includegraphics[width=0.33\textwidth]{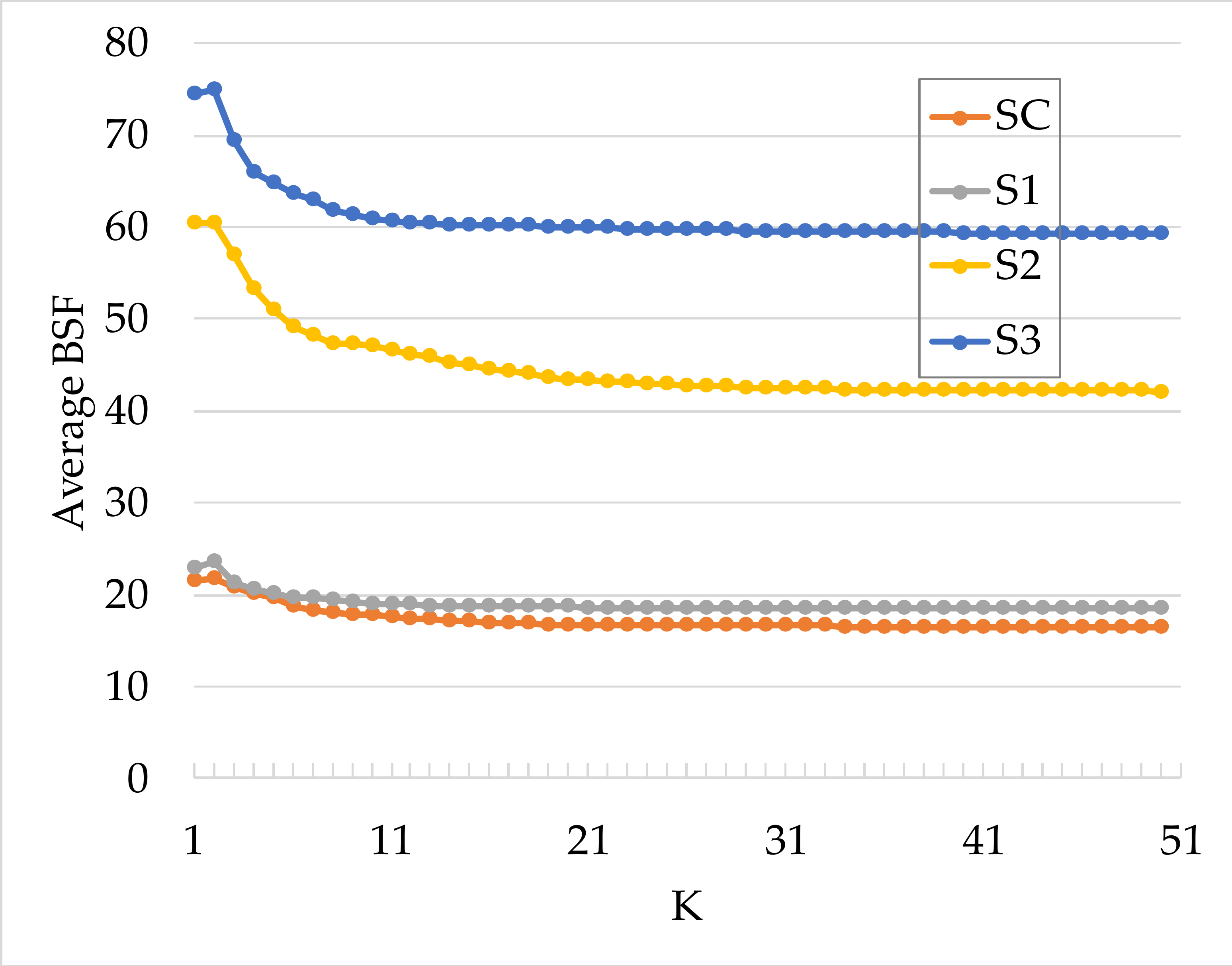}}
\subfloat[]{
\label{fig:K-AUPR}
\includegraphics[width=0.33\textwidth]{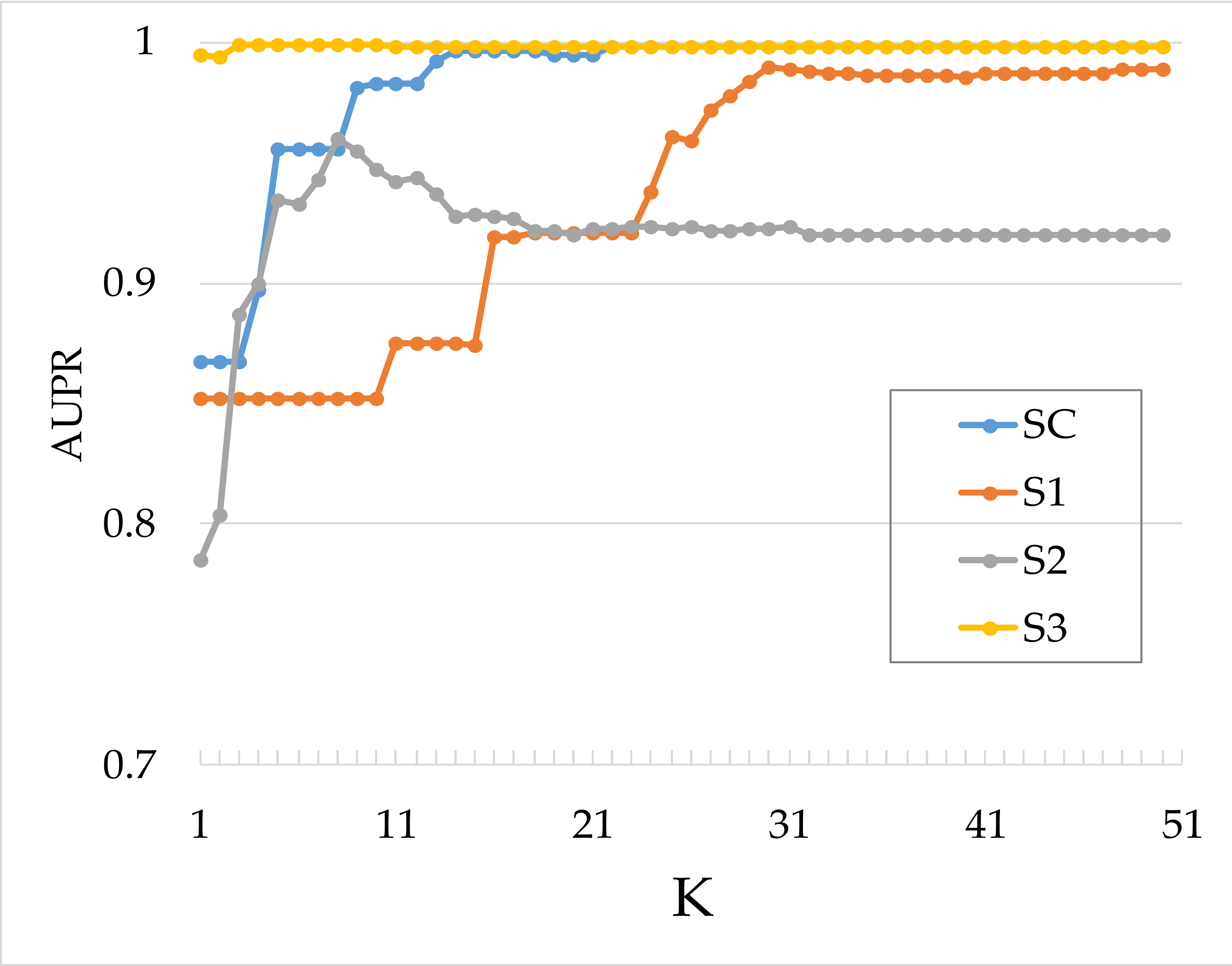}}
\caption{Evaluation results for the data sets using the proposed detection method with different values of the parameter $K$. (a) Average \emph{LCG}, (b) average \emph{BSF}, and (c) \emph{AUPR}. Best viewed in color.}
\label{fig:VaryK}
\end{figure*}

\begin{figure*} [htbp]
\centering 
\includegraphics[width=0.8\textwidth]{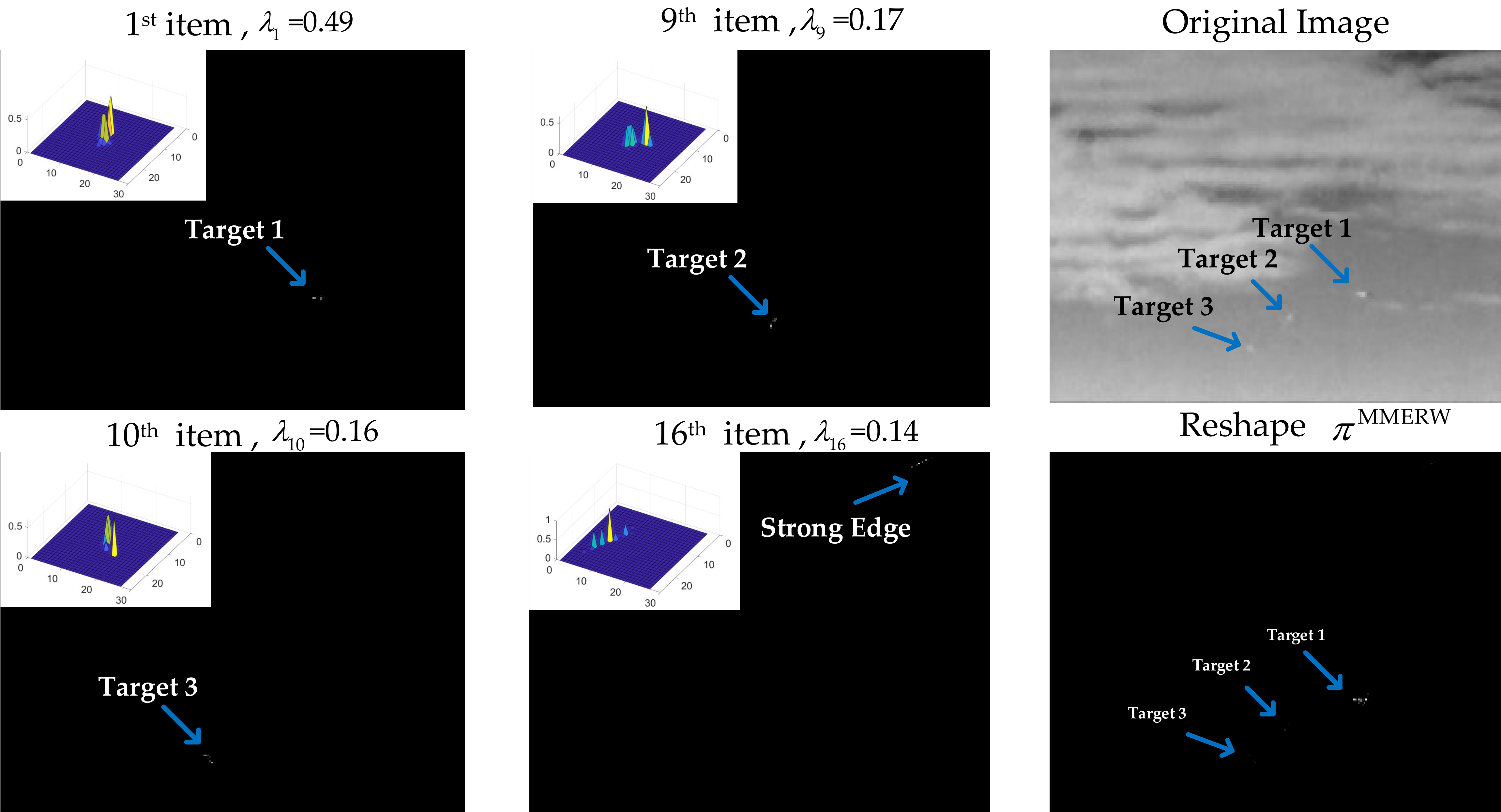}
\caption{2D visualization of some items composing the output stationary distribution of the proposed HMERW model. $\boldsymbol{\pi}^\text{HMERW}$ is computed according to \eqref{eq:HMERW stationary} by setting $K=30$. The upper left corner of the left four images displays 3D mesh of image patch indicated by blue arrows. Among the top 30 items, Target 2 is enhanced in the 9th, 13th and 28th items, Target 3 is enhanced in the 10th and 17th items, three kinds of strong edges are enhanced in the 16th, 18th and 19th items, and Target 1 is enhanced in the remaining items.}
\label{fig:2D-visualization}
\end{figure*}

\subsubsection{Analysis of $K$}\label{sec:analysis-K}
Recall \eqref{eq:HMERW stationary} that $K$ controls the ``accuracy'' of graph decomposition. On the one hand, large $K\in[1,N]$ provides rich information of the primal graph, but on the other hand, there may generate extra false detections if $K$ is too large. Meanwhile, larger $K$ requires more computation time. However, when there exists several targets in the infrared image, the detection model may miss the dim ones if $K$ is too small. Therefore, it is a trade-off to choose a suitable $K$ for our detection method. To see how the parameter $K$ affects the method, we vary $K$ from 1 to 50 with an interval of 1 and test it on the data sets.

Fig. \ref{fig:VaryK} presents the evaluation results of the proposed detection method with different $K$ for the data sets, from which we can see that the best choice of $K$ is different for different data sets. In detail, Fig. \ref{fig:K-LCG} presents a plot of the average \emph{LCG} results versus varying values of $K$ for the data sets, which evaluates the performance of target enhancement against the change of the parameter $K$. As shown in Fig. \ref{fig:K-LCG}, the results of \emph{LCG} for the data sets S2, S3 and SC increase as $K$ increases within 13, and remain approximately unchanged when $K$ exceeds 13. However, one may notice that \emph{LCG} slightly decreases when $K$ exceeds 30 for S3. This indicates that although larger $K$ can better emphasize characteristics of small targets, it may also mistakenly enhance the neighboring clutters around the targets if $K$ is too large.   For S1, the values of \emph{LCG} roughly increase as $K$ increases within a wider range $[1,50]$ (compared to the other data sets). This is because there are two dim small targets in the infrared images of S1, and they can only be enhanced when $K$ reaches a certain value. 

The indicator \emph{BSF} is used to evaluate the performance of background suppression, as defined in \eqref{eq:BSF}. Fig. \ref{fig:K-BSF} shows the average \emph{BSF} results for the data sets tested by the proposed method with different values of $K$. Notice that the values of \emph{BSF} for all the data sets slowly decreases with the increase of $K$. This is due to the effect of the parameter $K$ on the HMERW map. According to \eqref{eq:HMERW stationary}, although large $K$ can provide rich information for the model to enhance the characteristics of small targets, it also incorrectly emphasizes some undesired information of clutters, and thus reduce \emph{BSF}. 

Fig. \ref{fig:K-AUPR} displays the \emph{AUPR} results for different values of the parameter $K$, from which we can see that the parameter $K$ has different effects on different data sets. 
To be specific, S1 contains several infrared images with multiple small targets, so it requires large $K$ to search dim targets in more subspaces. Before $K$ reaches 30, the \emph{AUPR} roughly increases as $K$ increases. After $K$ exceeds 30, the \emph{AUPR} remains steady.
The image quality of S2 is quite poor, there exists many PNHBs in the images. Even worse, in some frames, the dim small target is blurred by heavy clouds and becomes even dimmer. However, by setting a moderate $K(=9)$, the proposed HMERW-based method can achieve a great improvement of \emph{AUPR} for S2 (improved by 0.17 compared with $K=1$). 
As for S3, in which a bird flies across buildings and the images are overexposed, the \emph{AUPR} result is not sensitive to the change of $K$.
Moreover, \emph{AUPR} reaches 1 when $K$ exceeds 10. 
SC is a collection of infrared images with small targets embedded in diverse backgrounds, it is found that our method can achieve perfect detection accuracy ($\emph{AUPR}=1$) for SC by setting $K>=11$. This demonstrates the robustness of the proposed method to various detection scenes. 
Note that the proposed HMERW model degenerates into the primal MERW when $K$ is set to 1. A promising finding is that the \emph{AUPR} results of setting $K>1$ are always better than those of setting $K=1$, from which we can conclude that the proposed HMERW is superior to the primal MERW for multiple small targets detection. In practical applications, setting $K$ to 30 can provide satisfactory detection performance.

\subsection{Analysis of multiple small target detection} As mentioned above, the proposed detection method based on the HMERW can achieve multiple small targets detection if we properly initialize the parameters. In the following, we perform experiments to analyze the mechanism of the proposed HMERW for detecting multiple small targets. As observed from \eqref{eq:HMERW stationary}, the stationary distribution $\boldsymbol{\pi}^\text{HMERW}$ of the HMERW model is comprised of normalized summation of $K$ items, each of which is computed by multiplying $\lambda_k$ and $(\boldsymbol{\psi}^{(k)})^2$, where $\lambda_k$ is one of the top $K$ eigenvalues of the weight matrix $\textbf{W}^\text{HMERW}$ and $\boldsymbol{\psi}^{(k)}$ is the corresponding eigenvector. 

To see how each item contributes to the output stationary distribution, Fig. \ref{fig:2D-visualization} presents two dimensional (2D) visualization of some items of $\boldsymbol{\pi}^\text{HMERW}$ by feeding a representative infrared image with multiple small targets to the HMERW. As shown in Fig. \ref{fig:2D-visualization}, there are three small targets in the original infrared image, two of which, i.e., Target 2 and Target 3, are so dim that they are difficult to be distinguished. Target 1, the most salient one of the targets, is well enhanced in most items, including the top 8 items of $\boldsymbol{\pi}^\text{HMERW}$. As for the dim ones, Target 2 is enhanced in the 9th,13th and 28th items, and Target 3 is enhanced in the 10th and 17th items. In some posterior items (16th, 18th and 19th items), some strong edges, such as the junctions of clouds and sky, are also mistakenly strengthened. However, the contribution of these posterior items to the output $\boldsymbol{\pi}^\text{HMERW}$ will be weakened due to their small eigenvalues. Moreover, these mistakenly enhanced strong edges will be further suppressed after combining $\boldsymbol{\pi}^\text{HMERW}$ and the cofficient vector \textbf{c} according to \eqref{eq:fusion-map}. This way, by selecting a suitable $K$ for the proposed HMERW-based method, we can achieve multiple small targets detection with only few false detections.

\begin{figure*} [htbp]
\centering 
\includegraphics[width=1\textwidth]{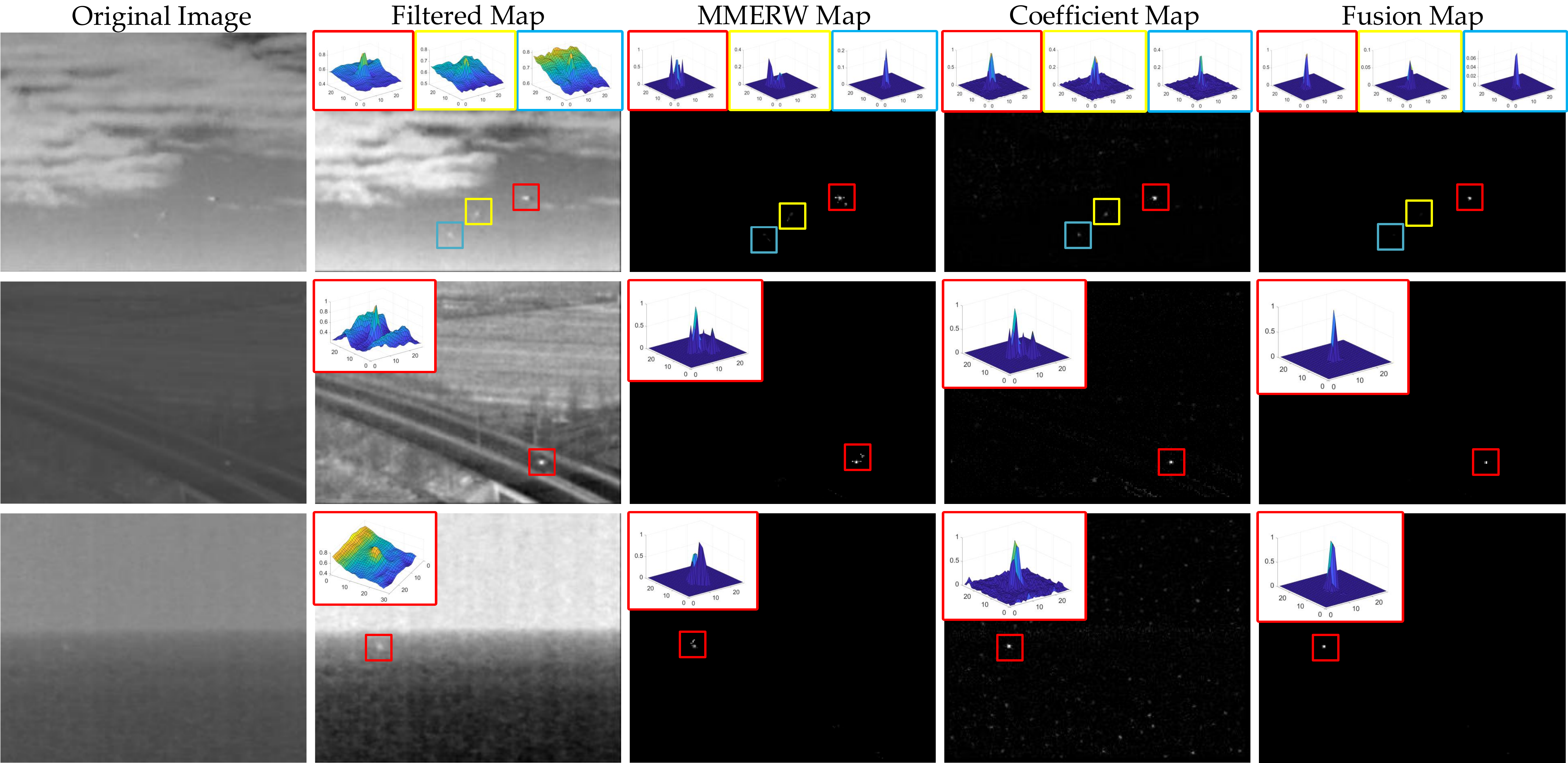}
\caption{Visual examples of the output of different components after processing the original infrared images by the proposed method. The zoom-in image patches placed in the corner are the 3D meshes of target regions (indicated by corresponding border color if there are multiple targets, best viewed in color).}
\label{fig:components}
\end{figure*}

\subsection{Analysis of Components}
The proposed HMERW-based small target detection method is comprised of two major components, that is, the HMERW map and the coefficient map. To validate the effectiveness of each component for target enhancement and background suppression, we display some visual examples of the output of different components in Figure \ref{fig:components}. 
It is clear that all the targets are well enhanced and their backgrounds are purely clean in the fusion map. This good performance benefits from the combination of the HMERW map (transformed from $\boldsymbol{\pi}^\text{HMERW}$) and the coefficient map. 
To be specific, most clutters are eliminated in the HMERW map. However, small targets, together with several nearby pseudo targets are enhanced, as shown in the zoom-in 3D meshes of the HMERW maps in Figure \ref{fig:components}.
These pseudo targets result from the transpose of weight matrix in \eqref{eq:symmetric-our-weight}, as discussed in \ref{sec:coefficient}. 
In contrary, in the coefficient map, these pseudo targets can be removed while strong clutters such as cloud edges, sea-sky boundary and road boundary remain. Therefore, these two components can complement each other w.r.t. small target enhancement and background suppression.

To quantitatively analyze the performance of different components, we present average \emph{LCG} and \emph{BSF} results for the data sets in Table \ref{tb:components}. Note that the HMERW map and the coefficient map are obtained by processing the filtered map. Compared with the filtered map, both the HMERW map and the coefficient map show great improvement in the results of \emph{LCG} and \emph{BSF}, which indicates that both the components can enhance small target and suppress background to different extents. Another finding from Table \ref{tb:components} is that the HMERW map always performs better in the \emph{BSF} results than the coefficient map for all the data sets, while the coefficient map achieves higher \emph{LCG} results. Accordingly, it can be deduced that the HMERW component predominates in clutters suppression, whereas the coefficient component contributes more to target enhancement. By taking the advantage of each component, the fusion map obtained by fusing the individual components gain highest results in the terms of \emph{LCG} and \emph{BSF} results. In summary, both the HMERW and coefficient components play important roles in target enhancement and background suppression, and their combination further improves the performance.

\begin{table}[htbp]
\centering
\caption{Average \emph{LCG} and \emph{BSF} results of different components for the data sets.}
\label{tb:components}
\begin{tabular}{l l l l l l}
\toprule
\multicolumn{2}{l}{} &Filtered &HMERW &Coefficient  &Fusion\\
\multicolumn{2}{l}{} &Map     &Map         &Map            &Map\\
\midrule
S1 &$\overline{\emph{LCG}}$  &0.64   &6.57 &7.95  &8.89\\
     &$\overline{\emph{BSF}}$  &0.95 &11.24 &9.44 &18.46\\
\midrule
S2 &$\overline{\emph{LCG}}$  &0.70 &7.25 &8.22 &8.75\\
     &$\overline{\emph{BSF}}$  &1.01 &24.95 &21.54 &42.55\\
\midrule
S3 &$\overline{\emph{LCG}}$  &0.88  &8.07 &8.70 &9.78\\
     &$\overline{\emph{BSF}}$  &1.01 &35.88 &37.60 &59.60\\
\midrule
SC   &$\overline{\emph{LCG}}$  &0.94 &7.03 &7.99 &8.93\\
       &$\overline{\emph{BSF}}$    &0.94 &10.13 &9.71 &16.72\\    
                                           
\bottomrule
\end{tabular}
\end{table}

\begin{table}[htbp]
\begin{threeparttable}
\centering
\caption{Indicator results of ablation experiments for data set S1.}
\label{tb:ablation}
\begin{tabular}{l l l l l l}
\toprule
&A&B&C&D&E\\
\midrule
&MERW                                          &HMERW                                                       &MERW 							&HMERW                                      &HMERW\\
&+${\textbf{W}^\text{E}}^{*}$     &+$\textbf{W}^\text{E}$                    &+${\textbf{W}^\text{H}}^{*}$      &+$\textbf{W}^\text{H}$    &+$\textbf{W}^\text{H}$\\
&                                                       &                                                      &      
&                                                     &+Fusion\\
\midrule
$\overline{\emph{LCG}}$  &1.13   &1.82   &3.74 &6.57  &8.89\\
\midrule
$\overline{\emph{BSF}}$   &5.21   &2.56 &17.66 &11.24  &18.46\\
\midrule
\emph{AUPR}                    &0.33   &0.47  &0.85 &0.92  &0.99\\                                     
\bottomrule
\end{tabular}
\begin{tablenotes}
\footnotesize
        \item[*] $\textbf{W}^\text{E}$ denotes $\textbf{W}^\text{Euclidean}$ and $\textbf{W}^\text{H}$ denotes $\textbf{W}^\text{HMERW}.$
\end{tablenotes}
\end{threeparttable}
\end{table}

\subsection{Ablation study}\label{sec:ablation}
To further validate the contributions of  the HMERW, designed weight matrix $\textbf{W}^\text{HMERW}$ and fusion technique to our detection method, we perform ablation experiments on data set S1 and list their results in Table \ref{tb:ablation}. Specifically, MERW is the primal maximal entropy random walks model (i.e., HMERW with $K=1$), HMERW is our modified version with $K=30$, $\textbf{W}^\text{E}$ denotes the Euclidean weight matrix of \eqref{eq:Euclidean-W}, $\textbf{W}^\text{H}$ denotes the designed weight matrix of \eqref{eq:symmetric-our-weight}, and \emph{Fusion} means fusing the stationary distribution map with a coefficient map according to \eqref{eq:fusion-map}. Here, we treat the combination of MERW + $\textbf{W}^\text{E}$ as a base model. For fair comparison, infrared images of S1 are filtered beforehand. 

According to Table \ref{tb:ablation}, we summarize several observations as follows.
\begin{enumerate}
\item Combination B (HMERW + $\textbf{W}^\text{E}$) shows better \emph{LCG} and \emph{AUPR} results than combination A (MERW + $\textbf{W}^\text{E}$).
\item Combination C (MERW + $\textbf{W}^\text{H}$) shows better \emph{LCG}, \emph{BSF} and \emph{AUPR} results than combination A.
\item Combination D (HMERW + $\textbf{W}^\text{H}$) shows better \emph{LCG} and \emph{AUPR} results than combinations B and C.
\item Combination E (HMERW + $\textbf{W}^\text{H}$ + Fusion) owns the best \emph{LCG}, \emph{BSF} and \emph{AUPR} results among all the combinations.
\end{enumerate}
Based on the above observations, several conclusions can be drawn as follows. 
\begin{enumerate}
\item Observation 1 illustrates that HMERW helps to detect multiple small targets in terms of small target enhancement and detection accuracy, compared with the primal MERW. Note that S1 is a data set containing multiple small targets in single frames. 
\item Observation 2 demonstrates the effectiveness of our designed weight matrix $\textbf{W}^\text{HMERW}$ in target enhancement, background suppression and detection accuracy.
\item Observation 3 validates that both the HMERW and our designed weight matrix contributes to target enhancement and multiple small targets detection. 
\item Observations 4 indicates the effectiveness of the fusion technique in target enhancement, background suppression and detection accuracy.
\item All the above observations demonstrates that each component plays important role in the proposed detection method and contributes to detecting multiple small targets.
\end{enumerate}

\subsection{Comparison to baseline methods}\label{sec:comparison}
To demonstrate the superiority of the proposed small target detection method for small target detection, several state-of-the-art methods are used for comparison. To be specific, small target detection methods based on the new top-hat transformation (THT) \cite{BAI20102145}, the multiscale relative local contrast measure (MRLCM) \cite{8289318}, the flux density and direction diversity in gradient vector field (GDD-MFD) \cite{8360016}, the non-convex rank approximation minimization joint $\ell_{2,1}$ norm (NRAM) \cite{rs10111821}, and the facet kernel and random walker (FKRW) \cite{8705367} are selected as baseline methods. Besides, to verify the effectiveness of our modification of the primal MERW, we substitute the primal MERW for HMERW in the proposed detection method and take it as a competitor.  For fair comparison, $2\times2$ mean filtering is applied to all the competitors, except for FKRW (due to its own preprocessing techniques). Meanwhile, the parameter initialization of the baseline methods is consistent with the original papers, and parameters of the proposed method are initialized as follows, $R=5$, $K=30$ and $\lambda=10$.

\begin{table*}
\centering
\caption{Average $\emph{LCG}$ and $\emph{BSF}$ of different methods for the data sets.}
\label{tb:enhancement}
\begin{threeparttable}
\begin{tabular}{l l l l l l l l l l}
\toprule
\multicolumn{2}{l}{}   &Filtered  &THT  &MRLCM    &GDD-MFD  &NRAM   &FKRW  &MERW  &HMERW\\
\midrule
S1 &$\overline{\emph{LCG}}$ &0.64               &$\textbf{6.39}_2$   &1.63           &2.37              &3.83       &5.66      &4.30       &$\textbf{8.89}_1$\\
     &$\overline{\emph{BSF}}$   &0.95               &2.41   &1.93           &3.65              &16.84     &12.65    &$\textbf{22.79}_1$     &$\textbf{18.46}_2$\\
\midrule
S2 &$\overline{\emph{LCG}}$  &0.70               &$\textbf{8.48}_2$   &1.56           &2.13              &0.85        &8.10     &5.89       &$\textbf{8.75}_1$\\
     &$\overline{\emph{BSF}}$  &1.01                 &18.66  &4.29           &7.84              &39.20      &31.34    &$\textbf{60.66}_1$&$\textbf{42.55}_2$\\
\midrule
S3 &$\overline{\emph{LCG}}$  &0.88                &9.04   &3.21           &1.60             &9.36       &6.12     &$\textbf{9.60}_2$       &$\textbf{9.78}_1$\\
      &$\overline{\emph{BSF}}$  &1.01                 &23.55 &8.55           &10.18           &33.22       &45.65    &$\textbf{74.27}_1$&$\textbf{59.60}_2$\\
\midrule
SC &$\overline{\emph{LCG}}$  &0.94                &$\textbf{8.15}_2$   &2.64           &1.58            &5.15        &7.65       &7.26       &$\textbf{8.93}_1$\\
      &$\overline{\emph{BSF}}$  &0.94                 &4.64    &3.27           &4.60            &16.30      &12.28     &$\textbf{21.32}_1$&$\textbf{16.72}_2$\\    
                                           
\bottomrule
\end{tabular}
\begin{tablenotes}
\footnotesize
        \item[*] The bolder data indicates the top two indicators, and the subscript indicates the top ranking. The "Filtered" maps are obtained by applying $2\times2$ mean filtering to the original images.
\end{tablenotes}
\end{threeparttable}
\end{table*}

\subsubsection{Target enhancement and background suppression}
The abilities of enhancing small targets and suppressing backgrounds are important for a small target detection method. The better the ability is, the easier the small target is detected. Table \ref{tb:enhancement} reports the average \emph{LCG} and \emph{BSF} results of testing different small target detection methods on the data sets. For the results of \emph{BSF}, our detection models (both the MERW-version and HMERW-version methods) perform better than the others.  This indicates that the two methods are better at suppressing background. In detail, the MERW-version method always achieves the optimal \emph{BSF} results and the HMERW-version method achieves the suboptimal results. This is due to the involvement of more sub-graphs in the HMERW, which has been analyzed in Section \ref{sec:analysis-K}. Another finding is that the HVS-based methods (MRLCM and GDD-MFD) obtain worse \emph{BSF} results than the other competitors, which implies that these method are poor at suppressing background.

In term of \emph{LCG}, the proposed HMERW-version method always performs best for all the data sets.
In addition, THT-based, FKRW-based and our MERW-version methods achieve fair performance of small target enhancement. 
By contrary, MRLCM-based and GDD-MFD-based methods show poor ability of target enhancement. The main reason is that these two methods are unable to enhance the characteristics of small targets buried in intricate clutters. 
The NRAM-based method obtains the suboptimal \emph{LCG} result for S3, however, it fails enhancing small targets for S2 (the images of which exist many PHNBs). This is because NRAM is sensitive to PNHBs, and it mistakenly enhances them (instead of the true small targets).

According to the above observations, the proposed small target detection method can always achieve top results on both \emph{LCG} and \emph{BSF} results. This demonstrates that our method is effective for enhancing small target and suppressing background.

\begin{figure*} [htbp]
\centering 
\includegraphics[width=1\textwidth]{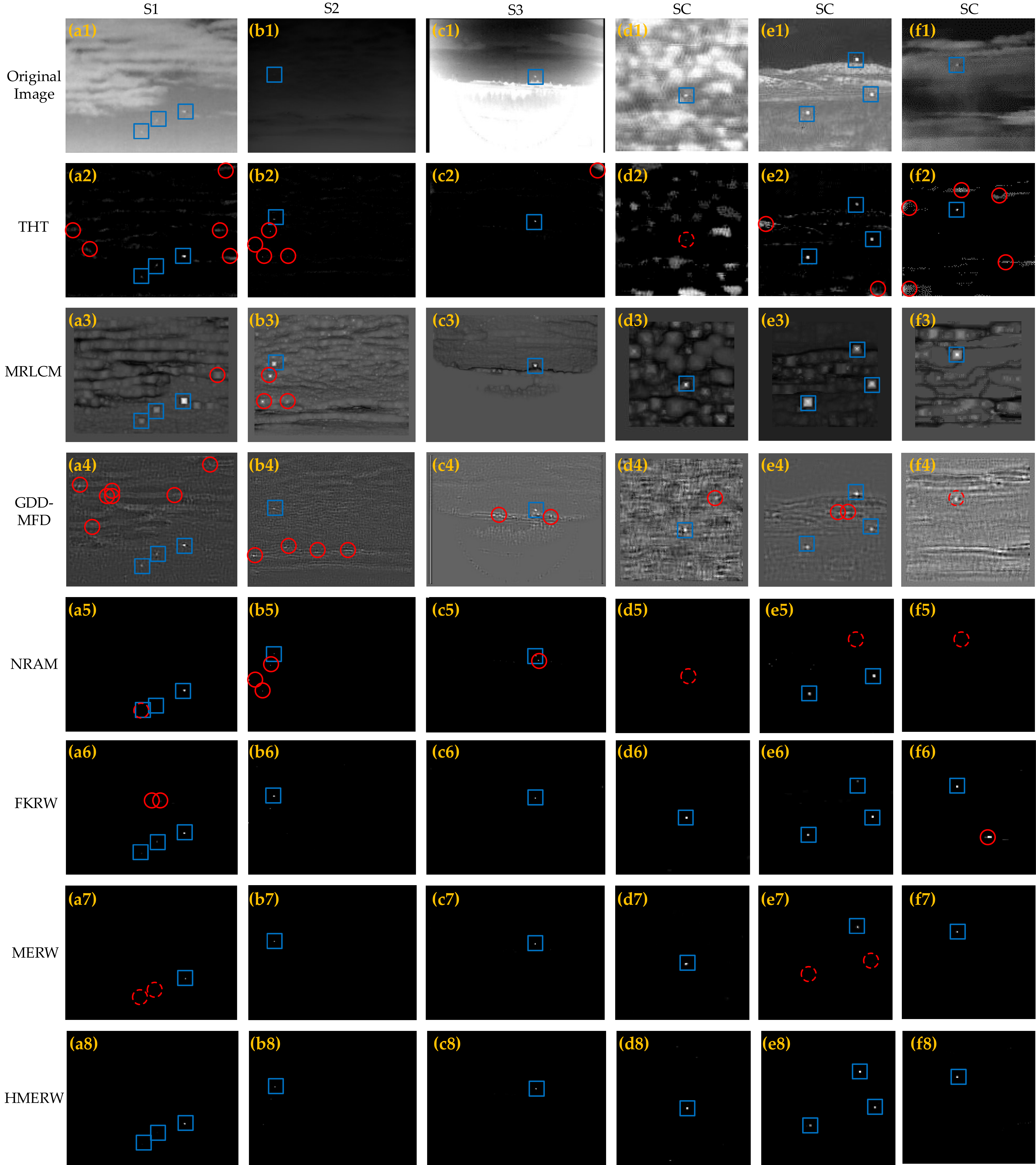}
\caption{Visual comparison of output maps of different small target detection methods for some representative infrared images. The blue rectangle indicates the ground truth or positive detection, the red circle indicates negative detection, the dashed red circle indicates failure in detection. Best viewed in color.}
\label{fig:visual-maps}
\end{figure*}

\begin{figure*}[htbp]
\centering 
\subfloat[]{
\label{fig:PR-S1}
\includegraphics[width=0.4\textwidth]{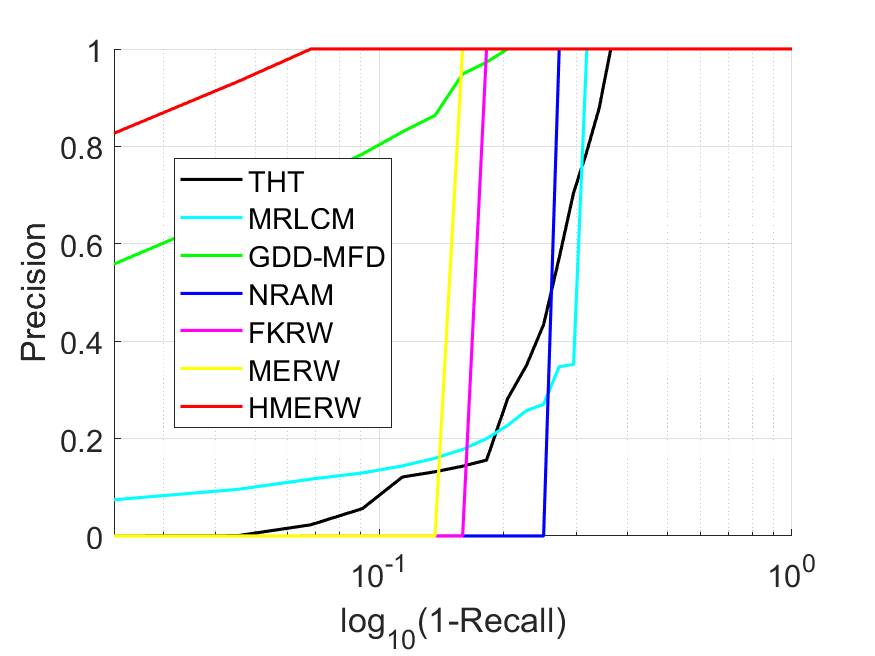}}
\subfloat[]{
\label{fig:PR-S2}
\includegraphics[width=0.4\textwidth]{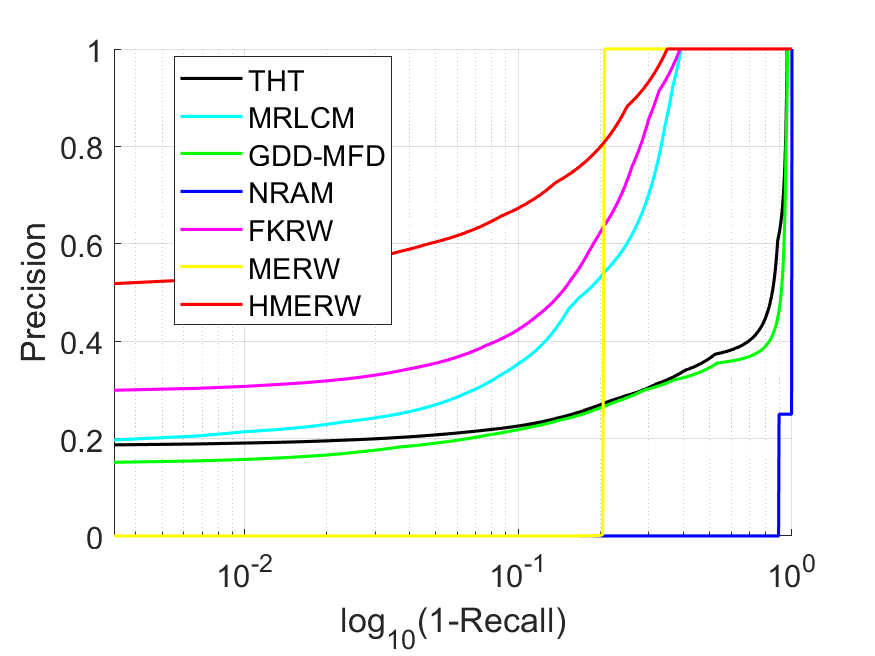}}\hfill
\subfloat[]{
\label{fig:PR-S3}
\includegraphics[width=0.4\textwidth]{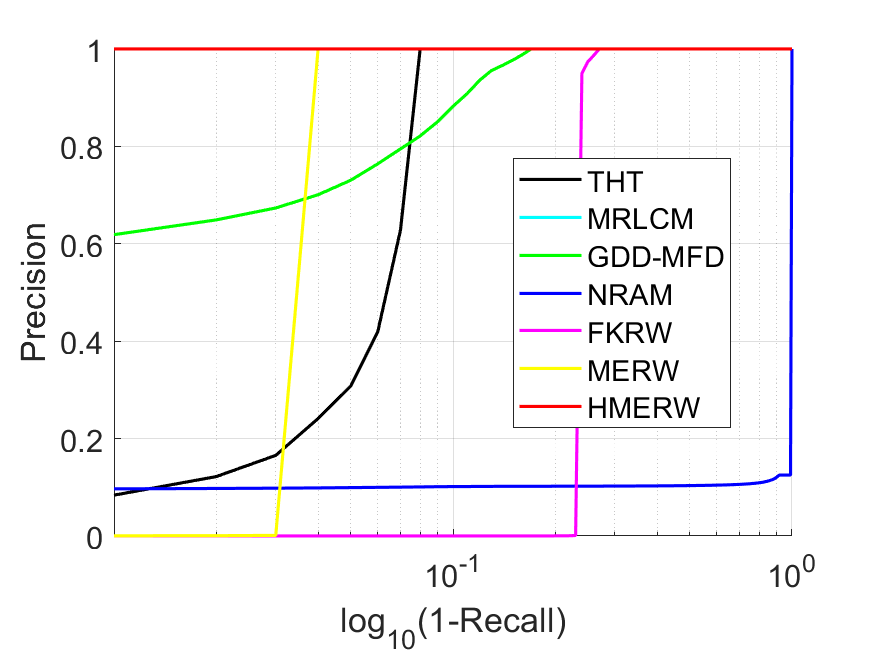}}
\subfloat[]{
\label{fig:PR-SC}
\includegraphics[width=0.4\textwidth]{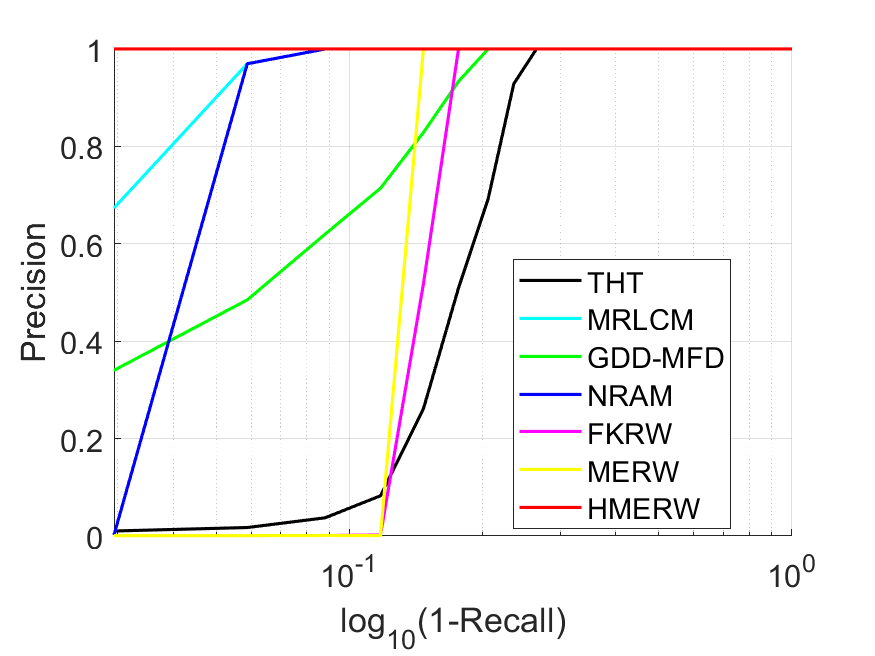}}
\caption{PR curves of testing different small target detection methods on the data sets \textbf{(a)} S1, \textbf{(b)} S2, \textbf{(c)} S3 and \textbf{(d)} SC. (For better visualization, the abscissa is set to $log_{10}(1-Recall)$.)}
\label{fig:PR}
\end{figure*}

\begin{table*}[htbp]
\centering
\caption{\emph{AUPR} results and running time of different small target detection methods for the data sets.}
\label{tb:AUPR}
\begin{tabular}{l l l l l l l l l}
\toprule
                & &THT  &MRLCM &GDD-MFD  &NRAM  &FKRW &MERW  &HMERW\\
\midrule
S1 &\emph{AUPR}  &0.75   &0.75        &0.95             &0.74       &0.83     &0.85       &$\textbf{0.99}$\\
     &time (s) &0.02  &9.18         &1.72             &1.14      &0.08      &1.58       &1.98\\ 
\midrule
S2 &\emph{AUPR}  &0.39   &0.82         &0.36            &0.03       &0.85     &0.80       &$\textbf{0.93}$\\
     &time (s) &0.03  &16.44       &2.61             &3.00       &0.12     &2.31       &2.79\\ 
\midrule
S3 &\emph{AUPR}  &0.95   &0.99        &0.97             &0.11        &0.76     &0.99       &$\textbf{1.00}$\\
     &time (s) &0.03   &13.12      &2.91             &2.68        &0.39     &2.11       &2.45\\
\midrule
SC  &\emph{AUPR} &0.82  &0.98        &0.93             &0.96        &0.85     &0.87         &$\textbf{1.00}$\\    
       &time (s) &0.02 &3.77        &0.98             &0.45        &0.09     &0.63        &0.73\\       
\bottomrule
\end{tabular}
\end{table*}

\subsubsection{Detection accuracy}
Fig. \ref{fig:visual-maps} displays visual comparison of output maps obtained by testing different methods on some practical examples. As seen from Fig. \ref{fig:visual-maps} (a2)-(f2), the THT-based method performs well in suppressing homogeneous clutters, while performs badly in suppressing inhomogeneous and complicated clutters. 
The local descriptors MRLCM (Fig. \ref{fig:visual-maps} (a3)-(f3)) and GDD-MFD (Fig. \ref{fig:visual-maps} (a4)-(f4)) show good ability to enhance small targets, however, they are unable to eliminate most clutters in the images. 
Compared with the THT, MRLCM and GDD-MFD, the NRAM-based (Fig. \ref{fig:visual-maps} (a5)-(f5)) method shows better performance in suppressing heavy clutters and strong interferences. However, it misses the detections of  a sailing boat on the sunlit lake (Fig. \ref{fig:visual-maps} (d5)) and an airplane flying across heavy clouds, which shows its poor robustness to various detection scenes. 
Moreover, it is unable to detect multiple small targets in a single frame. 
As shown in Fig. \ref{fig:visual-maps} (a5) and (e5), the NRAM-based method misses one of total three small targets in each frame. 
Likewise, our MERW-version method (Fig. \ref{fig:visual-maps} (a7)-(f7)) is good at suppressing clutters, while bad at detecting multiple small targets. 
Among the competitors, the FKRW-based and the proposed HMERW-version methods are the best two methods that can not only greatly enhance small targets and suppress background, but also detect multiple dim small targets in a single frame. 
In comparison to the FKRW-based method, the HMERW-version method is more powerful in suppressing pseudo targets. Comparing Fig. \ref{fig:visual-maps} (a6), (f6) and (a8), (f8), the HMERW-based method generates fewer false detections than the FKRW-based method under complicated scenes. 

For quantitative comparison, we test different methods on the data sets and obtain their PR curves and \emph{AUPR} results, as listed in Fig. \ref{fig:PR} and Table \ref{tb:AUPR}. 
A clear observation from Table \ref{tb:AUPR} is that the proposed HMERW-version method always achieves best \emph{AUPR} results for all the data sets, which demonstrates the superiority of our method in term of detection accuracy. 
For S1, the proposed HMERW-based method always owns the highest precision under the same recall. Specifically, as shown in Fig. \ref{fig:PR-S1}, our method can robustly detect 38 of total 44 targets ($Recall=0.86$) without generating any false detections. 
In contrast, except for the GDD-MFD-based method (reaching $Precision=0.86$), none of the baseline methods can reach $Recall=0.86$ with an acceptable detection precision (e.g., $Precision=0.3$). 
Further notice that two really dim small targets appear from the 24th frame of S1 (e.g., Target 2 and Target 3 of the 26th frame are shown in Fig. \ref{fig:2D-visualization}), and there are total 14 of them in S1. 
Note that our method can robustly detect 11 of them with only 3 false alarms (i.e., achieve $Recall=0.93$ with $Precision=0.93$), while the suboptimal  GDD-MFD-based method costs much more false alarms to detect these 11 dim targets (i.e., achieves $Recall=0.93$ with $Precision=0.73$). 
The above observations demonstrate that the proposed HMERW-version small target detection method is superior to the baseline methods in term of multiple small targets detection. 

S2 is quite a challenging data set for small target detection, in which small targets are very blurry and there exists many PNHBs. 
Despite the challenge, the proposed HMERW-version method still works and obtains the highest and satisfactory \emph{AUPR} result. The HMERW-version method obtains $\emph{AUPR}=0.92$ for S2 and reach $Recall=1$ with $Precision=0.51$.
The suboptimal \emph{AUPR} result is achieved by FKRW, it obtains $\emph{AUPR}=0.85$ and reach $Recall=1$ with $Precision=0.30$. 
On the contrary, the NRAM-based method performs worst among the competitors, due to its sensitiveness to PNHBs. 
An intriguing finding is that our MERW-version method surpasses all the other competitors and  maintains $Precision=1$ before $Recall$ reaches 0.80. However, its $Precision$ falls sharply to 0 after $Recall>0.80$, as shown in Fig. \ref{fig:PR-S2}. The reason for this phenomenon is as follows. In some frames of S2, small targets are blurred by heavy clouds, which results in their absences from the first-level sub-graph.
By comparison, the HMERW-version method can robustly detect these absent small targets in more sub-graphs. 
This further indicates the effectiveness of our modification of the primal MERW for detecting small targets submerged in intricate clutters.

For S3, the proposed HMERW-version method achieves the best detection performance ($\emph{AUPR}=1$), and the MRLCM-based method has the second best performance ($\emph{AUPR}=0.99$). 
On the other hand, the NRAM-based and the FKRW-based methods perform worse than the other methods, as shown in Fig. \ref{fig:PR-S3}. Note that infrared images of S3 are overexposed (an example is shown in Fig. \ref{fig:visual-maps} (c1)) and have dark margins. In such abominable condition, the NRAM-based and FKRW-based methods fail in detecting small targets. 

As for SC, a collection of single-frame images with small targets buried in different scenes, the proposed HMERW-version method obtains perfect detection performance with $\emph{AUPR}=1$. Conversely, the THT-based method performs worst among the competitors. This demonstrates the robustness of the proposed HMERW-version method to diverse detection scenarios.  

According to the above observations, we can draw a conclusion that the proposed HMERW-based small target detection method outperforms the baseline methods in term of detection accuracy and multiple small target detection.
\subsubsection{Computational complexity}
To show the computational complexity of the proposed method, we report average running time of different methods for each data set in Table \ref{tb:AUPR}. Among the competitors, the THT-based method is the most efficient method, while the MRLCM-based method is the most time-consuming one. It can be seen that our MERW-version and HMERW-version methods take about the same running time as the GDD-MFD-based and NRAM-based methods. Another finding from Table \ref{tb:AUPR} is that the HMERW-based method spends about $15\%$ running time on implementing matrix decomposition and spends most of time on constructing the weight matrix. Fortunately, as shown in Algorithm \ref{alg:framework}, the construction of the weight matrix is very suitable for parallel processing. Therefore, the proposed method can be easily accelerated by using a field programmable gate array (FPGA) to achieve real-time detection.

\section{CONCLUSION}\label{sec:conclusion}
In this paper, we propose a single-frame small target detection algorithm based on a hierarchical maximal entropy random walk (HMERW) model. First, we analyze the primal MERW in describing global uniqueness and its limitation for practical applications. Next, we develop HMERW based on a proposed graph decomposition theory to alleviate MERW's limitation. Then, we further design a specific weight matrix for HMERW to incorporate both global and local characteristics of a small target. In addition, a coefficient map is constructed based on the specially designed weight matrix,  in which small targets are well enhanced while the clutters are suppressed. After that, a fusion map is built by fusing the output map of the HMERW and the coefficient map. Extensive experiments have demonstrated the effectiveness of the proposed method for small target enhancement, background suppression and high detection accuracy. Moreover, it is verified that the proposed detection algorithm is better at detecting multiple small targets in a single frame, compared with the state-of-the-art methods.
\ifCLASSOPTIONcaptionsoff
  \newpage
\fi



\bibliographystyle{IEEEtran}
\bibliography{IEEEabrv,bare_jrnl}
\end{document}